\documentclass[twocolumn,showpacs,preprintnumbers,amsmath,amssymb]{revtex4}

\usepackage{amsmath,amsfonts,amssymb}
\usepackage{graphicx}
\usepackage{setspace}
\usepackage{tocloft}
\usepackage{mathtools}
\usepackage{dcolumn}
\usepackage{bm}

\begin{document}

\preprint{APS/123-QED}
\title{Application of Superpixels to Segment Several Landmarks in Running Rodents}

\author{Omid Haji Maghsoudi}
 \email{o.maghsoudi@temple.edu}
\author{Annie Vahedipour}
\author{Benjamin Robertson}
\author{Andrew Spence}
 \homepage{http://www.spencelab.com/~Andrew Spence}

\affiliation{%
Spence Lab., Bioengineering, College of Engineering, Temple University, 12th Street, Philadelphia, PA, USA, 19122\\
}%

        
\begin{abstract}
Examining locomotion has improved our basic understanding of motor control and aided in treating motor impairment. Mice and rats are the model system of choice for basic neuroscience studies of human disease. High frame rates are needed to quantify the kinematics of running rodents, due to their high stride frequency. Manual tracking, especially for multiple body landmarks, becomes extremely time-consuming. To overcome these limitations, we proposed the use of superpixels based image segmentation as superpixels utilized both spatial and color information for segmentation. We segmented some parts of body and tested the success of segmentation as a function of color space and SLIC segment size. We used a simple merging function to connect the segmented regions considered as neighbor and having the same intensity value range. In addition, 28 features were extracted, and t-SNE was used to demonstrate how much the methods are capable to differentiate the regions. Finally, we compared the segmented regions to a manually outlined region. The results showed for segmentation, using the RGB image was slightly better compared to the hue channel. For merging and classification, however, the hue representation was better as it captures the relevant color information in a single channel.
\end{abstract}

\pacs{Valid PACS appear here}
\maketitle
\section{Introduction}
\label{sect:intro}
Understanding how animals, including humans, move is a grand challenge for modern science that has direct impact on our health and wellbeing. It is a useful instrument with which to explain the biological world, and to treat human and animal disease. It most directly impacts the treatment of musculoskeletal injuries {\cite{Arnold92}} and neurological disorders {\cite{Cirak11}}, can improve prosthetic limb design {\cite{Herr12}}, and aids in the construction of legged robots {\cite{Alphadog12}}. \\
One of the main features of locomotion is the gait (relative timing of leg recirculation: e.g., walk, run, trot, or gallop).  How gait is chosen and the regulation of gait can provide detailed information about the condition of a subject {\cite{Clarke99}}. Although significant insight into the neuromechanical basis of movement has been gained {\cite{Orlovsky99}}, there are many questions to be asked in this area; such as: how does gait control reflect the morphology and dynamics of the fast moving body? How is sensory feedback used during fast legged locomotion? \\
New genetic tools such as optogenetics and chemogenetics are making possible unprecedented manipulations of the nervous system in intact, freely behaving mice and rats. These include temporally fast manipulations, and therefore high frame rate kinematic data from these animals are increasingly important {\cite{courtine2008recovery}}.\\
Segmentation of body parts, including ear, nose, tail, and skin can provide valuable information to study biomechanics or the progress of diseases affecting motor controls or nervous systems. These points can be used, for example, to estimate the global position and orientation of the body, as well as the posture, of rodents {\cite{migliaccio11, baker05}}.\\
Rodents, especially mice and rats, are premier models of human disease and increasingly the model system of choice for basic neuroscience. High frame rates ($\geq$ 150 Hz) are needed to quantify the kinematics of running rodents, due to their high stride frequency (up to 10 Hz). Achieving an adequate number of strides to capture inter-stride variability may require 3-seconds or more of video; at least 450 frames need to be captured. This number increases rapidly with frame rate, which may be increased to capture sudden movements or reaction to impulsive perturbations, or with duration, which may be increased to yield large data sets for more sophisticated analyses of locomotor dynamics {\cite{Revzen12}}. Larger datasets are increasingly yielding insight {\cite{Wiltschko15}}, but cause difficulties in requiring bandwidth and space to store this data, algorithms to automatically track the desired animal body regions, and the required processing power and time to analyze them. The usual method to track the markers is manual clicking, simple thresholding, cross correlation, or template matching {\cite{hedrick2008software}}, which can be prohibitively time consuming for high frame rates and multiple views. Thresholding has been a popular method for segmentation and tracking of insect {\cite{noldus2002computerised}}; although, it cannot be used for tracking of an object showing variation in the intensity level amongst the frames. \\
Tracking of tip of paw is useful for many studies in biology, biomechanics, and robotics {\cite{wenger2016spatiotemporal}}. To automatically or semi-automatically track rodents paws, that can provide the required information for gait analysis, several methods have been proposed, including commercially available systems (Digigait {\cite{Dorman14, Gadalla14}}, Motorater, Noldus Catwalk {\cite{Huehnchen13, Hamers04, Parvathy13}}). These systems can be prohibitively expensive, and may only provide information about paws during the stance phase. In both research and commercial systems, tracking rodents has frequently relied on shaving fur and then drawing markers on the skin for subsequent tracking from raw video {\cite{Maghsoudi15}}, or on the attachment of retroreflective markers, and the use of optical motion capture systems. These methods have the drawback of requiring anesthesia and multiple handlings applications of markers, and the problem of animal removing the attached markers. \\
In addition to tracking of paws, tracking of joints, like nose, tail, ear, and all other parts of rodent body (referred to as skin here), provides information about kinematics of the running animal on a treadmill including pitch, roll, and yaw. However, tracking the whole body has been presented using thresholding and active contours {\cite{Spence13}}, simple kmeans and particle filters {\cite{gonccalves07}}, and auto-adjustable observation model enhanced particle filter results {\cite{pistori10}}; no method has been proposed to that specifically aims to track paws, nose, ear, tail, or skin (that with future work could be correlated to joint locations and potentially angles) from side view cameras.\\
There is a large amount of literature on automatic superpixel algorithms, for example, normalized cuts {\cite{Ren03}}, mean shift algorithm {\cite{Comaniciu02}}, graph-based method {\cite{Felzenszwalb04}}, Turbopixels {\cite{Levinshtein09}}, SLIC superpixels {\cite{Achanta12}}, and optimization-based superpixels {\cite{Veksler10}}. \\
However, superpixel methods have not been used for animal tracking and segmentation of different landmarks in the bodies of animal, there have been some studies for hand segmentation, tracking, and gesture recognition {\cite{shu2013improving, serra2013hand, li2013model, baraldi2014gesture, li2013pixel}}.\\
\begin{figure}[t!p]
     \centering
      \includegraphics[scale=0.37]{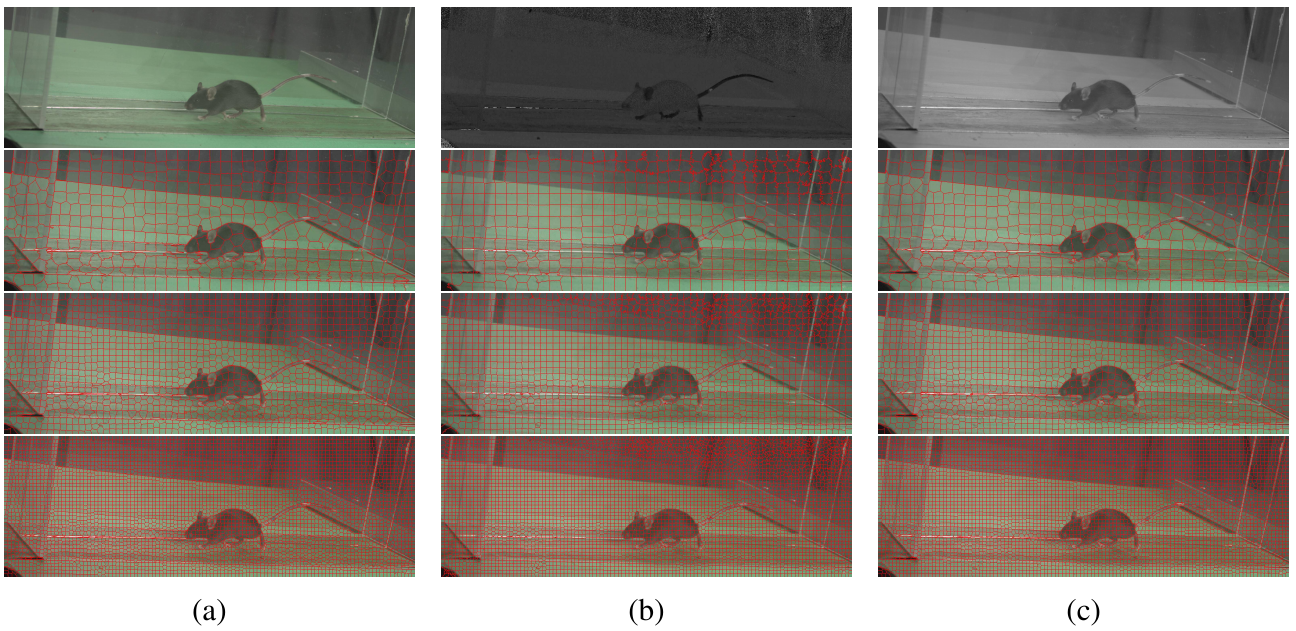}
\caption{A sample mouse frame in different color channels from different color spaces. a, b, and c show respectively the RGB image, hue channel from HSV color space, and the gray scale image. From top to bottom, the rows respectively show the original frames, SLIC segmentation results by 500 segments, 1500 segments, and 4500 segments.} \label{fig1}
\end{figure}
A superpixel based bag-of-words (BoW) approach was used to segment people walking in a parking lot {\cite{shu2013improving}}. Conditional random field (CRF) along with BoW model were used to differentiate the object from background. The proposed method by Smith et al. generated an output of the exact object regions instead of the bounding boxes generated by the previous methods. \\
Another method used superpixels a Random Forest to classify the superpixels generated from captured frames from human hand {\cite{serra2013hand}}. Then, a gesture recognition method based on exemplar SVMs was used to find the corresponding gesture for the frame. In another work {\cite{li2013pixel}}, superpixel based method was presented to segment hand. After applying the superpixel method on an image, texture and color features were extracted using Gabor filter, histograms of oriented gradients (HOG), binary robust independent elementary features (BRIEF), and oriented fast rotated BRIEF (ORB). Then, t-distributed stochastic neighbor embedding (t-SNE) was used to demonstrate the best features distinguishing the background from hand.\\
\begin{figure}[b!p]
\centering
      \includegraphics[scale=0.37]{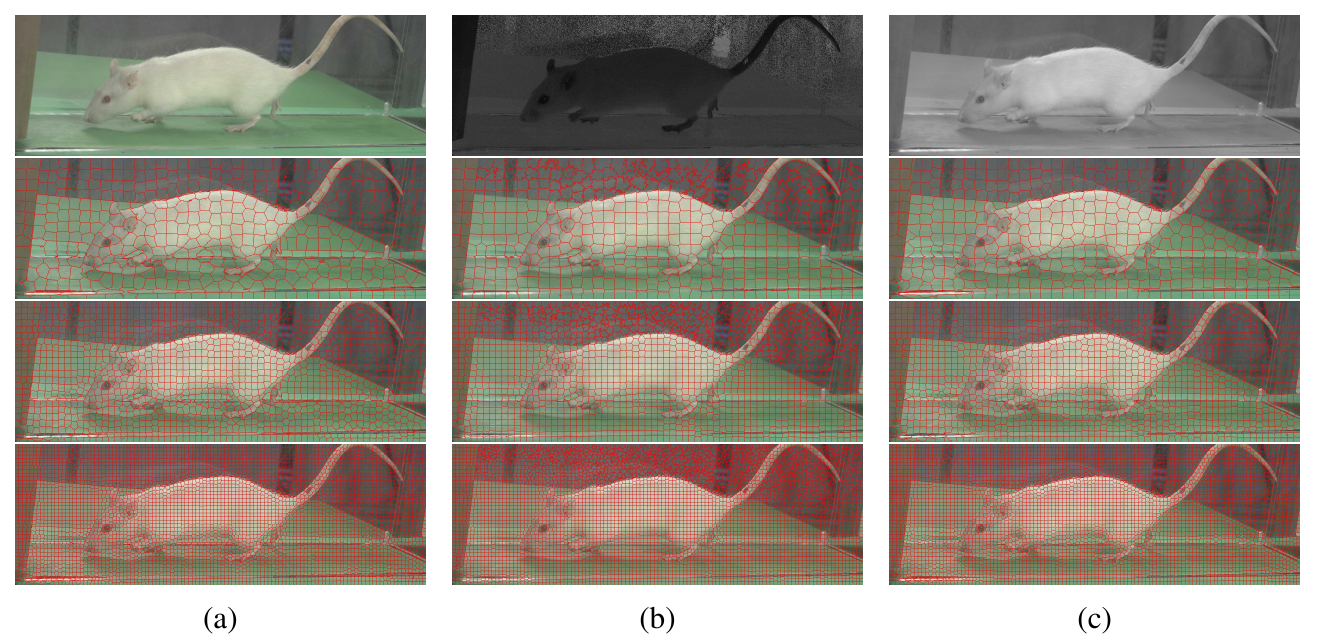}
\caption{A sample rat frame in different color channels from different color spaces. a, b, and c show respectively the RGB image, hue channel from HSV color space, and the gray scale image. From top to bottom, the rows respectively show the original frames, SLIC segmentation results by 500 segments, 1500 segments, and 4500 segments.} \label{fig2}
\end{figure}
Methods to detect the global behavioral state using thresholding {\cite{noldus2001ethovision}} have been used widely for behavioral experiments; but these methods cannot provide the required information for biomechanincs and related neuroscience applications as tracking of specific parts of body is needed. The main contribution of this study is to investigate the efficiency of a superpixel based method to segment these parts of body. We study the performance of simple linear iterative clustering (SLIC), graph based (Gb) {\cite{Felzenszwalb04}}, and quick shift (QS){\cite{vedaldi2008quick}} superpixels methods on RGB, hue channel from the HSV color space {\cite{Maghsoudi16_2}}, and the gray scale images. To determine the separability of our segmented regions, we extracted 28 features and applied t-SNE. In addition, we propose a tracking system to show the abilities of the discussed methods for biomechanics applications.
\begin{figure*}[t!p]
\centering
\includegraphics[scale=0.7]{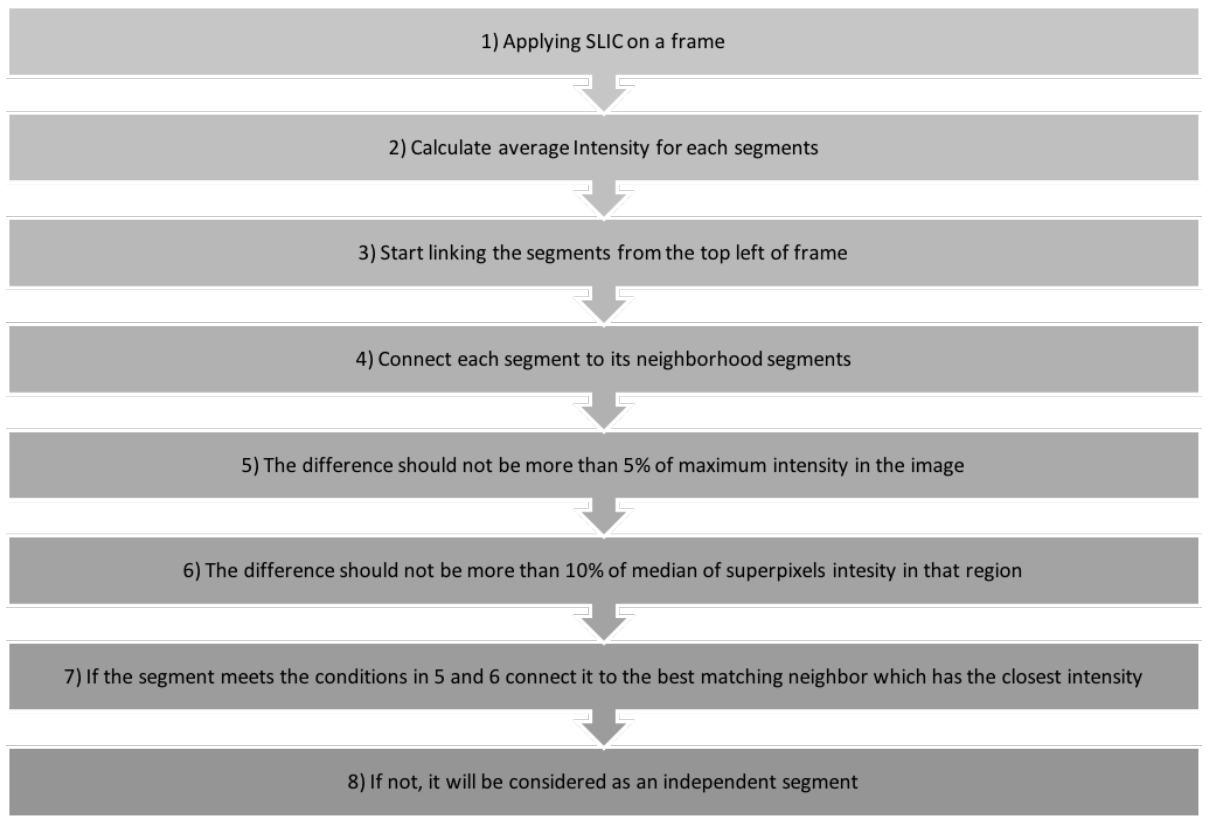}
\caption{This graph shows the algorithm used to connect the segments to each other after superpixel segmentations.} \label{fig3}
\end{figure*}
\section{The Treadmill System}
\label{sect:tmill} 
\subsection{Animals}
We analyzed data from female C57BL/6 mice and Sprague-dawley rats, because they are the most widely used strains in basic research and biomedicine. Animals were housed under a 12-12 hours light-dark cycle in a temperature-controlled environment with food and water available ad libitum. Animal procedures were approved by the Temple University Institutional Animal Care and Use Committee.
\subsection{Camera}
Ximea USB3 (Serial number: MQ022CG-CM) cameras were used to capture frames using a 250 Hz external synchronization signal. The trigger signal was generated and synchronized with a host PC using the triggerbox tools generously made available by the Straw Laboratory {\cite{Straw11, StrawGit}}. Briefly, the trigger pulses were generated by an Arduino Uno, running the triggerbox firmware. The Arduino was controlled via serial over USB by a standard desktop PC. The camera resolution was set to $2048\times700$ pixels at 8 bits depth, using a Bayer filter pattern to recover color.
\subsection{Treadmill and Tracking System}
We used a closed-loop treadmill system described in {\cite{Spence13}} to control and adjust the speed of treadmill while the mouse was running. The feedback loop helped us to keep animal in a specific place on treadmill (for example in middle) or control the speed of treadmill at specific speed while the animal was running in that specific region. We captured 1000 frames for each trial (providing four seconds of running).\\
Five cameras (one at top and four side-views) were used to capture the locomotion of the animal on the belt. An additional camera located at top of belt tracked the animal to provide the real-time feed for the visual servo-ing of the treadmill belt. Here, we analysed the frames captured just by one of the side views, the front left view of the mouse {\cite{Maghsoudi16}}.\\
We applied a control law to the treadmill belt speed that sought to keep the mouse at the mid-point of the belt. We further used the real-time feed of mouse position on the belt to apply a mechanical perturbation (a sudden vertical displacement of the belt surface, caused by an actuated camera under the belt) and captured two seconds before and after the perturbation applied.
\section{Superpixel}
\label{sect:superpixel}
The superpixel algorithm contracts and groups uniform pixels in an image. It has been widely used in many computer vision applications such as image segmentation and object recognition {\cite{Mori04, Li12}}. The superpixel concept was originally presented by Ren and Malik {\cite{Ren03}} as defining the perceptually uniform regions using the normalized cuts algorithm. The main merit of superpixel is to provide a more natural and perceptually meaningful representation of the input image. Therefore, compared to the traditional pixel representation of the image, the superpixel representation greatly reduces the number of image primitives and improves the representative efficiency. Furthermore, it is more convenient and effective to compute the region based visual features by superpixel, which has been shown to provide important benefits for vision tasks such as object recognition {\cite{Mori04}} or hand gesture recognition {\cite{shu2013improving, serra2013hand, li2013model, baraldi2014gesture, li2013pixel}}. \\
Here, we use SLIC superpixels segmentation on different color images. SLIC is a form of kmeans clustering for superpixels generation having two main advantages: the number of distance calculations is decreased by superpixel size and a weighted distance measure combines color and spatial relation which updates the size and compactness of superpixels. \\
The key parameter for SLIC is size of superpixels. First, $N$ centers are defined as cluster centers. Then, to avoid having centers that are on the edge of an object, the center is transferred to the lowest gradient position in a $3\times3$ neighborhood. The next step is clustering, as each of the pixels are associated with the nearest cluster center based on color information. It means that two coordinate components ($x$ and $y$) depict the location of the segment and three components (for example in the RGB color space, $R$, $G$, and $B$) are derived from color channels. SLIC finds and minimizes a distance (an Euclidean norm on 5D spaces) function defined as follow:
\begin{equation}
\label{eq:1}
D_{c} = \sqrt{(R_{j}-R_{i})^{2}+(G_{j}-G_{i})^{2}+(B_{j}-B_{i})^{2}},
\end{equation}
\begin{equation}
\label{eq:2}
D_{p} = \sqrt{(x_{j}-x_{i})^{2}+(y_{j}-y_{i})^{2}},
\end{equation}
\begin{equation}
\label{eq:3}
D = \sqrt{(\frac{D_{c}}{N_{c}})^{2}+(\frac{D_{p}}{N_{p}})^{2}}.
\end{equation}
Where $N_{c}$ and $N_{p}$ are respectively maximum distances within a cluster used to normalize the color and spatial proximity. Then, SLIC merges the pixels based on the calculated number to create superpixels. It should be said that SLIC is also constrained to ensure that the region does not grow more than twice the cluster radius; therefore, SLIC size plays an important role on how the segmentation is performed.
\section{Color Spaces}
\label{sect:col} 
The frames captured by cameras were in bayered raw images. We first converted them to the RGB color space using a debayering process {\cite{Maghsoudi16}}. The color information in the RGB color space is shared between all three channels of red, green, and blue. \\
The rodents' body carries different color information compared to the belt and the background in the frames. Therefore, we intend to use different color spaces to find the best color space for using superpixel segmentation. We tried to use A and B channels from the LAB color space showing Chroma information in image {\cite{Phung05}}; but, the intensity values were almost too close and did not provide enough distinctive information for SLIC segmentation. We were, however, able to use this segmentation approach on RGB, HSV, and gray scale images.  Sample images and segmentation can be seen in Figures \ref{fig1} and \ref{fig2} for mice and rats respectively.
\begin{figure*}[t!p]
                 \centering
                 \includegraphics[scale=0.4]{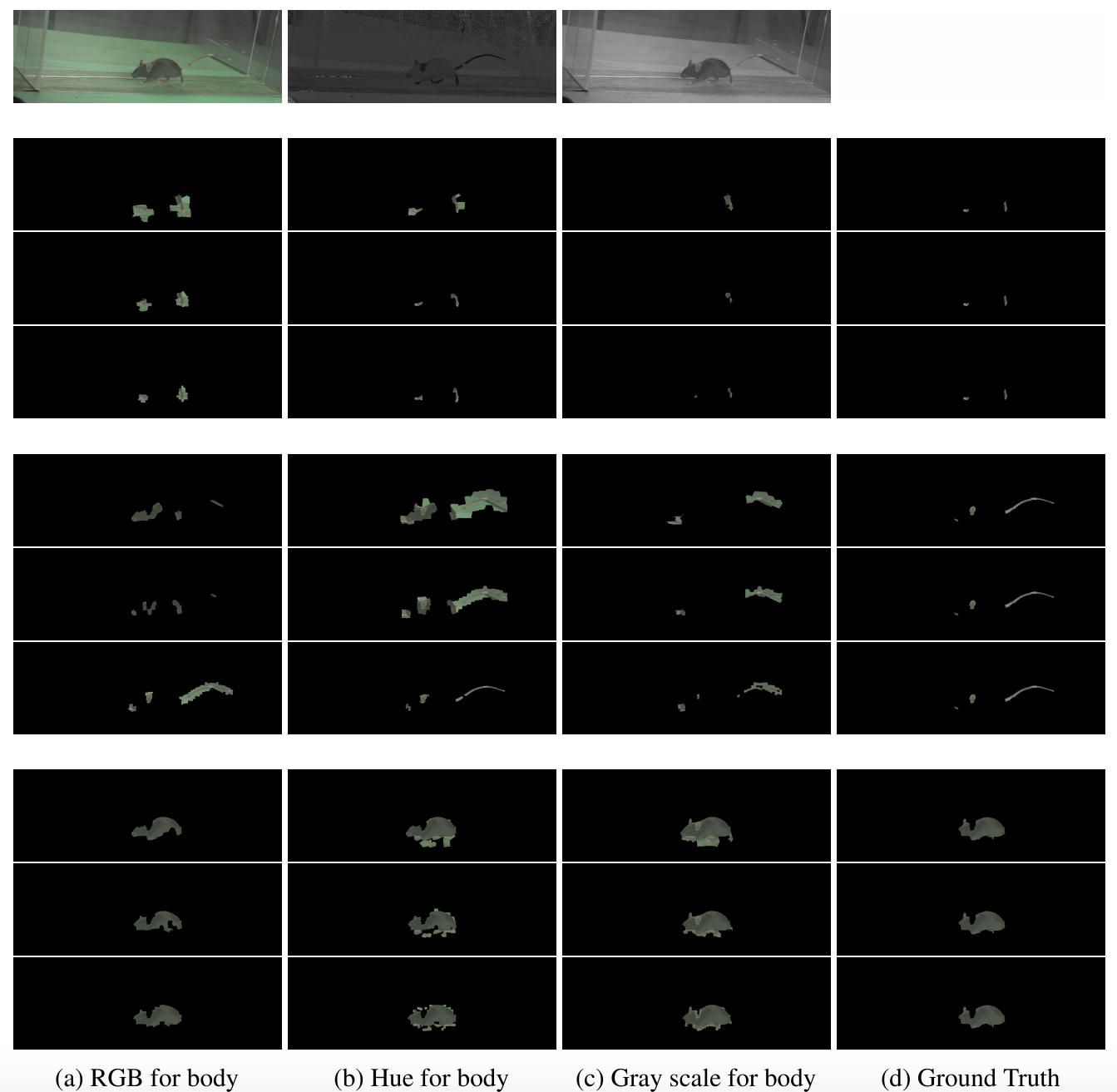}
\caption{A sample mouse frame in different color channels from different color spaces. The RGB image, hue channel from the HSV color space, gray scale image, and ground truth segmentation are illustrated respectively from left to right. From top to bottom, each of three rows respectively shows SLIC segmentation results by 500, 1500, and 4500 segments. The results illustrate three groups: paw segmentation, tail segmentation, and body segmentation.} \label{fig4}
\end{figure*}
\section{Segmentation}
\label{sect:seg}
Here, we investigate the SLIC superpixel segmentation method {\cite{Achanta12}} for different parts of body. We categorize the segmentation to three sections: paw segmentation; ear, nose, and tail segmentation; and skin (considered as body segmentation class in this paper) segmentation. After applying SLIC segmentation, a merging function was used to connect the neighbor superpixels. First, the center of superpixels were found by the following equation:
\begin{equation}
\label{eq:4}
C_{X_{SP_{i}}} = \dfrac{\Sigma X_{SP_{i}}}{\Sigma{i}}, C_{Y_{SP_{i}}} = \dfrac{\Sigma Y_{SP_{i}}}{\Sigma{i}}
\end{equation}
where X, Y, i, and C respectively show the horizontal coordinate, vertical coordinate, superpixel number after segmentation (between 1 and N which N is superpixel size), and center of a superpixel. Then, we calculated the average of image intensity from the channel (hue or gray scale) or channels (average of three channels of R, G, and B) by the following equation:
\begin{equation}
\label{eq:5}
I_{C_{SP_{i}}} = \dfrac{\Sigma I_{SP_{i}}}{\Sigma{i}}
\end{equation}
where I shows the average intensity for superpixels number i. Each superpixels was connected to the neighboring superpixels which had the closest average intensity expect the difference of this average was more than ten percent of image intensity or five percent should not be more than ten percent of the maximum intensity of image and ten percent of the median intensities in the connected region. This is achieved by finding the all superpixels ($SP_{i}$) having borders with a superpixel ($SP_{j}$) as follow: 
\begin{equation}
\label{eq:6}
A_{SP_{j}} = \lbrace (\forall SP_{i}) : SP_{i} \ has \ border \ with \ SP_{j} \rbrace
\end{equation}
where A is the set of superpixels considered as the neighbors of superpixels number j. This leads to find the intensity difference using:
\begin{equation}
\label{eq:7}
\small \begin{array}{   l   l    }
D_{SP_{k},SP_{j}} = \lbrace (\vert I_{C_{SP_{k}}}-I_{C_{SP_{j}}} \vert) : SP_{k} \in A_{SP_{j}}, \\
\vert I_{C_{SP_{k}}}-I_{C_{SP_{j}}} \vert < 0.10*I_{C_{SP_{j}}} \cap \vert I_{C_{SP_{k}}}-I_{C_{SP_{j}}} \vert < M \rbrace
\end{array}
\end{equation}
where D shows the difference intensity, and M is five percent of maximum intensity value (180 for hue channel and 255 for red, green, blue, and gray scale). Finally, the superpixels were connected to each other by the following equation:
\begin{equation}
\label{eq:8}
\begin{array}{   l   l    }
L_{SP_{k}} =  \lbrace (j) : \forall j, D_{SP_{k},SP_{j}} \neq \varnothing \rbrace \\
G = \lbrace SP{i}; \forall i \in L_{SP_{k}} \rbrace
\end{array}
\end{equation}
L and G respectively represent the indexes of all linked superpixels and the grouped superpixels. The segmentation algorithm is simplified and illustrated in Figure \ref{fig3}. \\
Regarding the importance of paws for biomechanics studies and the size of body compared with other landmarks, we report the results based on the importance for three manually classified merged regions: paw; skin (also referred as body); and ear, nose, and tail. 
\begin{figure*}[tp]
      \centering
      \includegraphics[scale=0.4]{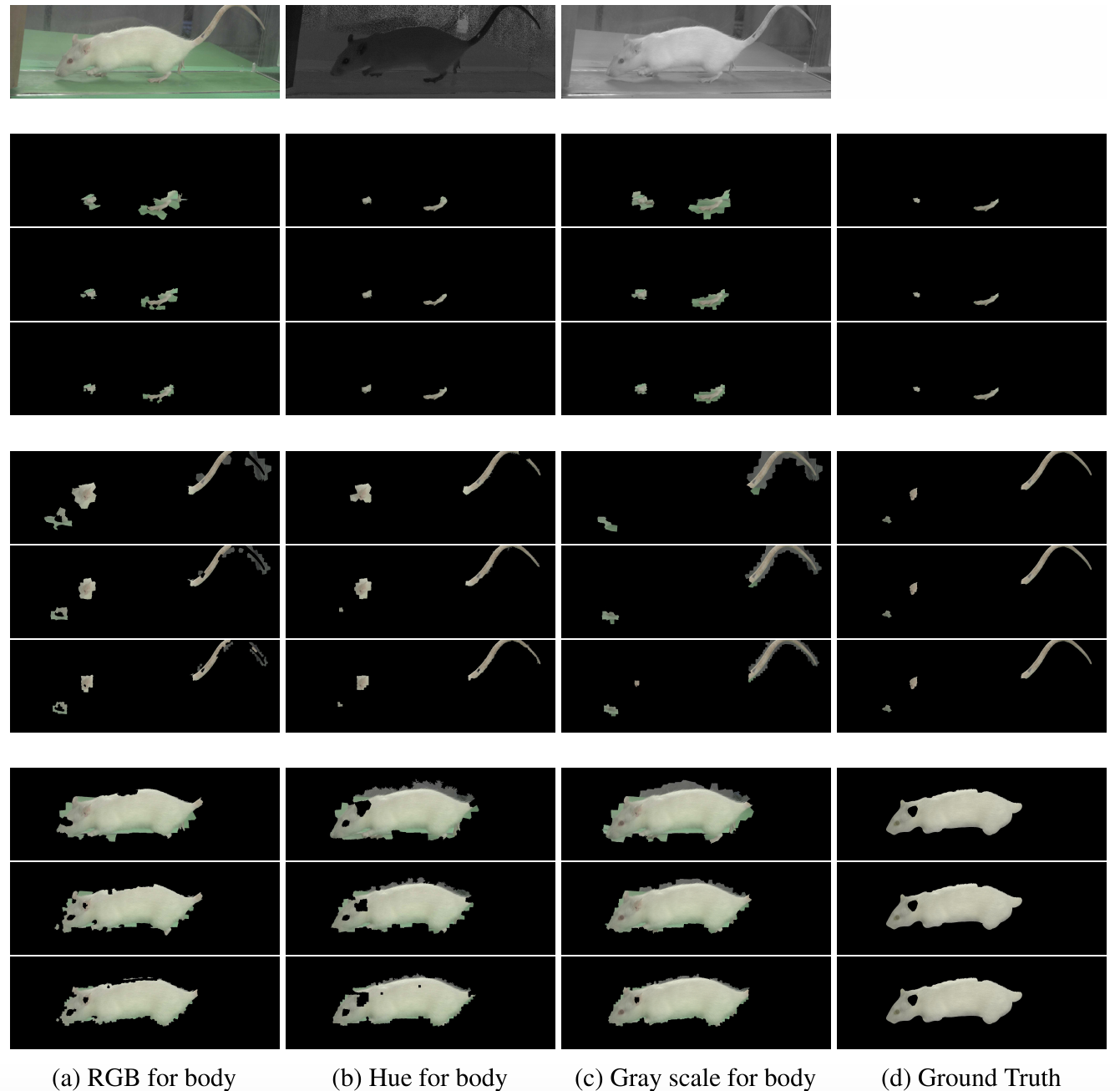}
\caption{A sample rat frame in different color channels from different color spaces. The RGB image, hue channel from the HSV color space, gray scale image, and ground truth segmentation are illustrated respectively from left to right. From top to bottom, each of three rows respectively shows SLIC segmentation results by 500, 1500, and 4500 segments. The results illustrate three groups: paw segmentation, tail segmentation, and body segmentation.} \label{fig5}
\end{figure*}
\subsection{Paw Segmentation}
The location of a foot is frequently one the most interesting regions of body for biology, biomechanics, and robotics; in our images, it can consist of 100 to 3500 pixels depending on the front and hind limbs, the camera positioning, stride cycle, and the mouse movement direction on treadmill. The shape has lots of changes especially on swing phase of stride cycle. Having variable shape, size, and position makes the paw segmentation difficult. There are, however two features that can be used to segment the paws: first, features derived from color and gray scale images, and second, texture features which are unique for paws. Here, we use superpixels for segmentation that mainly relies on the first feature. The segmentation using SLIC is shown in Figure \ref{fig1} for mice and in Figure \ref{fig2} for rats. 
\subsection{Ear, Nose, and Tail Segmentation}
Ear, nose, and tail (considered as tail segmentation class) are three parts of body that carry different color information than the skin. Despite lots of shape, size, and position variations for paws, the ear, nose, and base of the tail however are most closely coupled to movements of the center of the body/center of mass. Although, the tail moves with more variation (especially in terms of position), the base of the tail can be considered moving with the center of body, especially at high speeds. 
\subsection{Skin Segmentation for 3D Modeling}
Subtracting the paws, nose, ear, and tail leaves the body in the frames. The idea behind superpixels is to create meaningful "superpixels" that are collections of pixels with similar color information. The segmentation of skin as some meaningful pixels (superpixels) is an important step towards creating a 3D model of a mouse body using four views {\cite{Maghsoudi15}}.
\section{Features}
\label{sect:fet}
Two sets of features were extracted from the superpixels: texture and color features. The color features were the average of intensity for each of the superpixels and from four color channels, gray scale, green, saturation, and hue. This provided four features. The texture features were extracted by cropping the superpixel regions and calculating the co-occurrence matrix {\cite{albregtsen2008statistical}} on four different angles (0, 45, 90, and 145 degree). Then, following six features were extracted for each of the angles: contrast, dissimilarity, homogeneity, angular second moment, energy, and correlation {\cite{albregtsen2008statistical}}.
\begin{figure*}[t!p]
         \centering
      \includegraphics[scale=0.9]{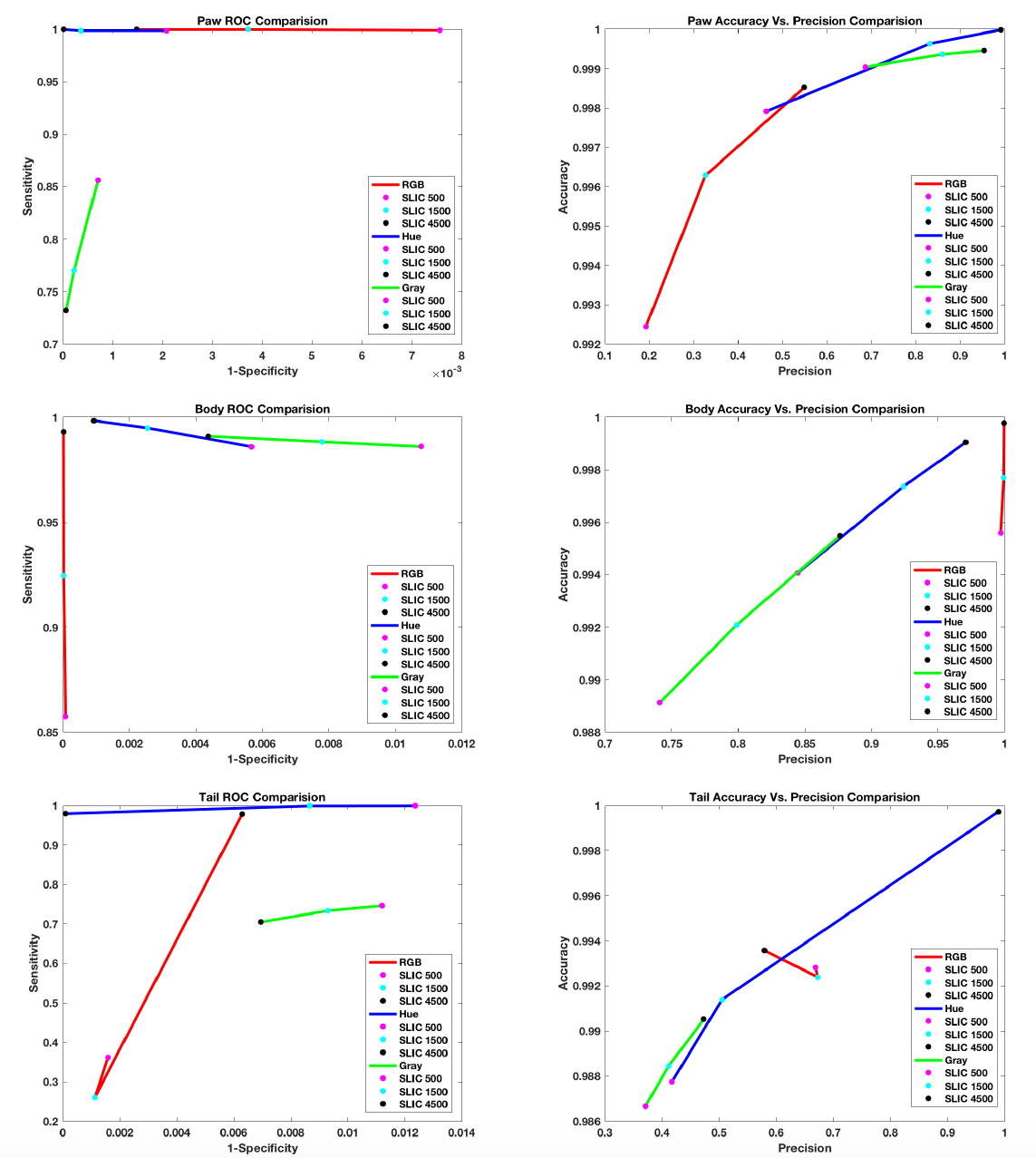}
\caption{Receiver operating characteristic (ROC) plot and accuracy versus precision plot for mice. The left graphs show the ROC plots and the right graphs illustrate the accuracy versus precision plots. The First, second, and third row of graphs show respectively the results related to segmentation of paws, body, and tail. Inside each of the graphs, the red, blue, and red green lines illustrate the results for RGB, hue, and gray scale images. In addition, three points on each of the line by magenta, cyan, and black colors show the results related to 500, 1500, and 4500 superpixels.}  \label{fig6}
\end{figure*}
\section{Tracker}
\label{sect:tra}
\begin{figure*}[t!p]
         \centering
      \includegraphics[scale=0.9]{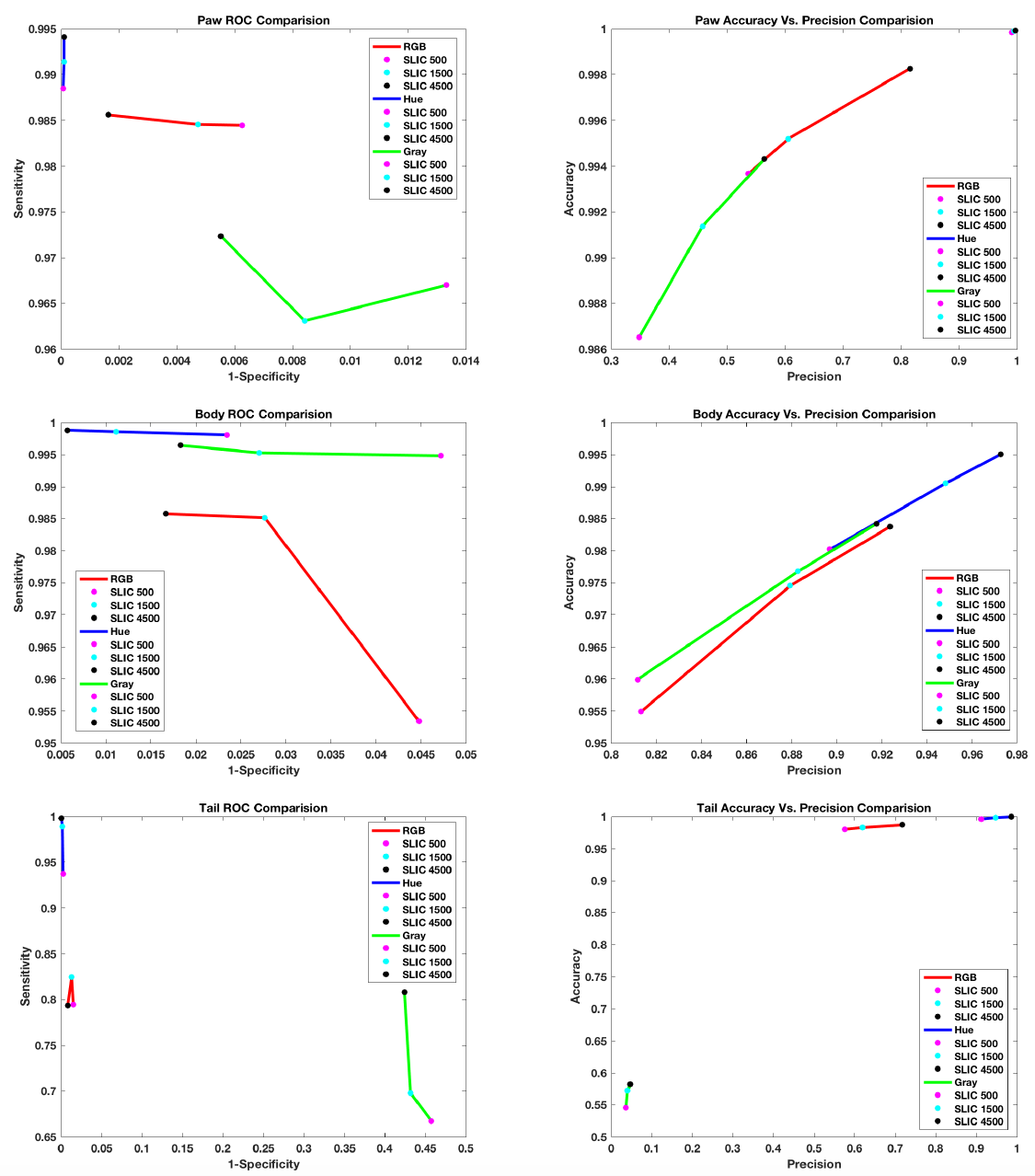}
\caption{Receiver operating characteristic (ROC) plot and accuracy versus precision plot for rats. The left graphs show the ROC plots and the right graphs illustrate the accuracy versus precision plots. The First, second, and third row of graphs show respectively the results related to segmentation of paws, body, and tail. Inside each of the graphs, the red, blue, and red green lines illustrate the results for RGB, hue, and gray scale images. In addition, three points on each line by magenta, cyan, and black colors show the results related to 500, 1500, and 4500 superpixels.}  \label{fig7}
\end{figure*}
After segmentation and merging of the superpixels using one of the alternate methods, SLIC, Gb, and QS, we use a tracker algorithm that is based on position, speed, size, and color information of the tracked region in the previous frame. A user was asked to click on the correct landmark on the first frame. We subsequently focused on an $80\times80$ pixel region of interest (ROI) given the user initialization in the first frame, because frame-to-frame landmark movement was always within this ROI, and considering only this ROI drastically reduces computation time. The size of image was selected based on the maximum displacement of center of body in rats (30 pixels). Then, we designed a function, referred to as the "tracker function", to assign a weight to each of objects remaining after segmentation. This function found the closest object to the previous tracked marker position, average of hue, size, and following the same speed and direction of movement. The object with the maximum value of this function was chosen as the tracked object in the current frame.\\
The tracker function can be simplified as follow:
\begin{equation}
W(k,f) = \left\{
\begin{array}{   l   l    }
closest \ object \ to \ T(f-1) \\
\ \ \ \ \ \ \ \ \ \ \ \ W(k,f) = W(k,f)+3 \\ 
moving \ in \ same \ direction \ T(f-1) \\
\ \ \ \ \ \ \ \ \ \ \ \ W(k,f) = W(k,f)+2 \\
minimum(abs(H(T(f-1)- H(M(k,f)))) \\
\ \ \ \ \ \ \ \ \ \ \ \ W(k,f) = W(k,f)+2 \\
minimum(abs(S(T(f-1)- S(M(k,f)))) \\
\ \ \ \ \ \ \ \ \ \ \ \ W(k,f) = W(k,f)+1 \\
minimum(abs(G(T(f-1)- G(M(k,f))))  \\
\ \ \ \ \ \ \ \ \ \ \ \ W(k,f) = W(k,f)+1
\end{array}
\right.
\end{equation} 
\begin{equation}
T(f) = M(f,where \ W(k,f) =Maximum(W(k,f)))
\end{equation} 
where W, T, and G are respectively the weighted function chosen based experiments, the tracked marker for the current frame, and average of gray scale image. k and f are respectively the superpixel number and the frame number.
\section{Results}
\label{sect:res}
\begin{figure}[t!p]
         \centering
      \includegraphics[scale=0.8]{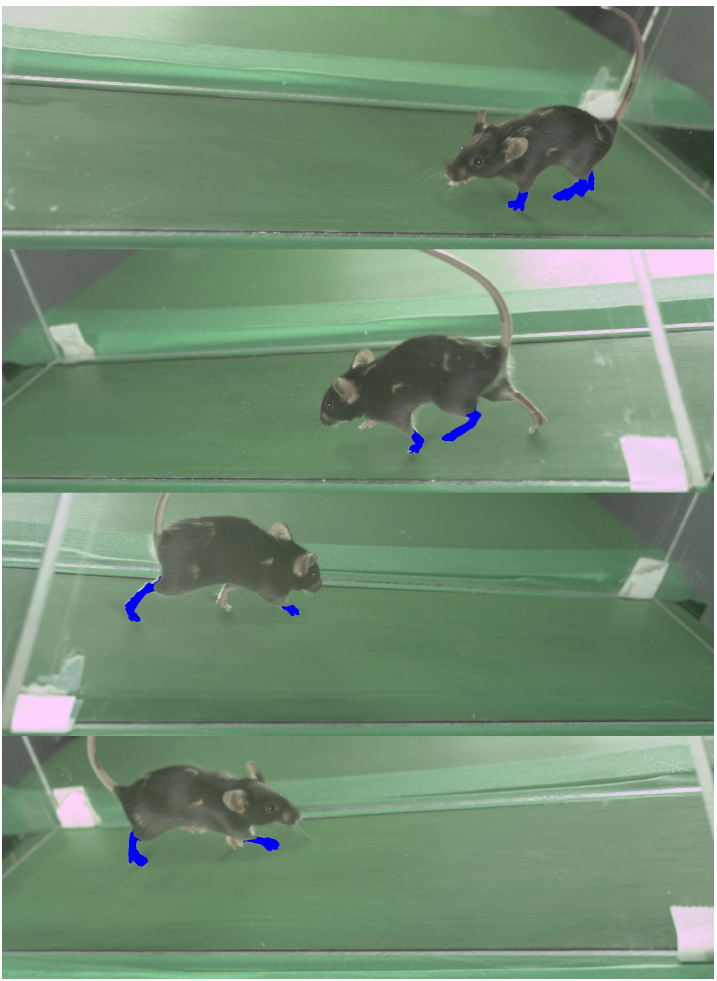}
\caption{A sample video showing the segmeneted and tracked paws by applying SLIC on the RGB image and having 1500 superpixels for 50 consecutive frames from four cameras (MP4, 7.5 MB).}  \label{fig8}
\end{figure}

\begin{figure}[t!p]
         \centering
      \includegraphics[scale=0.4]{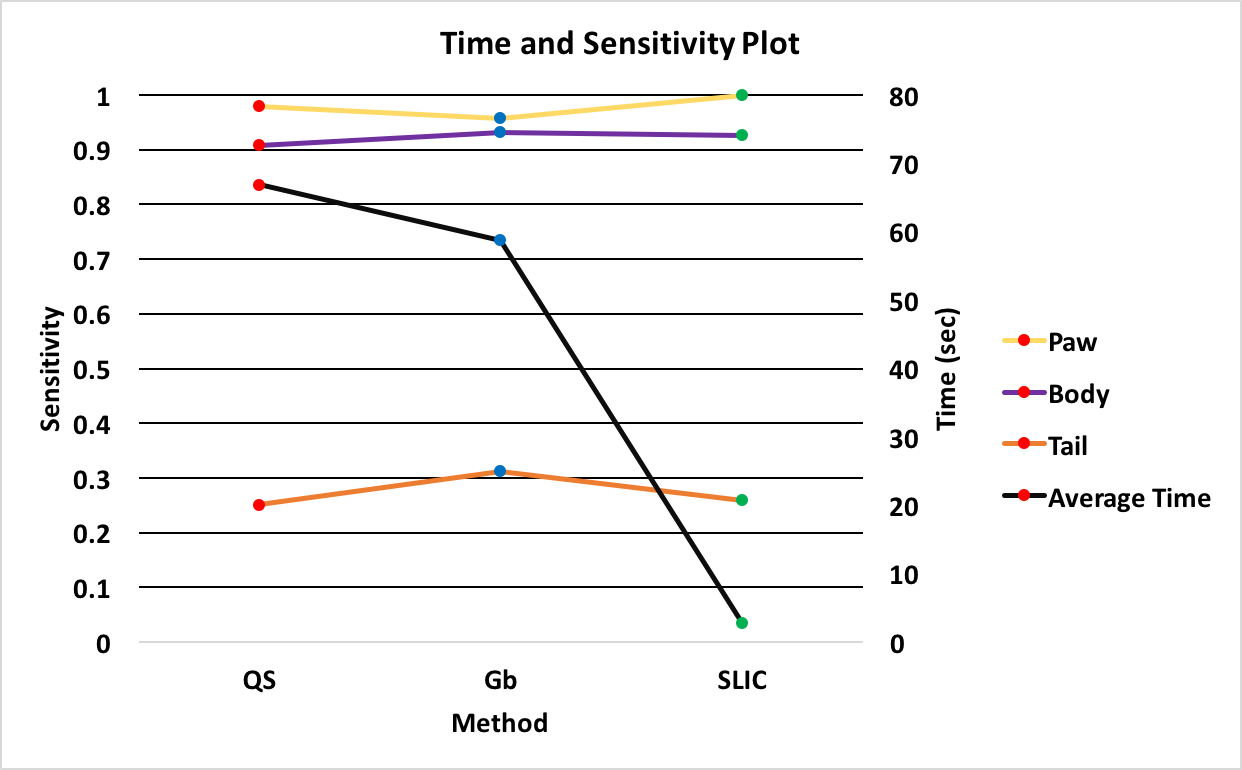}
\caption{The average time required to process the frames for the three superpixels methods and comparison between these methods using the sensitivity plot. The horizontal axis shows the methods. Left and right vertical axis respectively demonstrate the sensitivity and time results. The black graph shows the average time required to process the methods and the other three graphs show the sensitivity results to detect paw, body, and tail classes. The RGB image with 1500 superpixels was used to compare the results of the three methods.}  \label{fig9}
\end{figure}

To evaluate the segmentation, we used the frames captured from five mice and five rats. Two trials from each animal were selected just from the front right camera. Each trial created 1000 frames, but to test the method for different animals and reduce the manual burden of segmentation, we randomly selected 25 frames from each trial. Therefore, 250 frames from five mice and 250 frames from five rats were established as the database for this study. \\
The SLIC superpixels method was applied on three image types (RGB, hue channel, and gray scale) and at three different superpixels sizes: 500, 1500, and 4500. These numbers were selected based on the size of paw in the frames which can vary between 100 to 3500 pixels. The image size is $2048 \times 700$ which creates 1,433,600 pixels. SLIC method can generate superpixels that are twice or half initially specified size. This means that by specifying a superpixel size of 4500, we can have between 150 to 600 pixels in each of the superpixels ($1,433,600/4500 \simeq 300$).  Figure \ref{fig1} and Figure \ref{fig2} illustrate respectively how the SLIC is applied on a mouse sample frame and a rat sample frame.\\
Then, the process described in Figure \ref{fig3} was applied on the segmented regions to connect them to each other and create paws, nose, ear, tail, and skin. Figure \ref{fig4} and Figure \ref{fig5} show the segmented area using this method. To quantify the segmentation method, we needed to compare with a ground truth segmentation. The ground truth segmentation was done by manual supervision using a designed graphical interface in Matlab. This was then compared to the segmented regions using SLIC segmentation and our merging function. To do this comparison, we used the following measures:
\begin{equation} 
\label{eq:4}
\begin{aligned}
Sensitivity = \dfrac{TP}{TP+FN},\\
Specificity = \dfrac{TN}{TN+FP},\\
Precision = \dfrac{TP}{TP+FP},\\
Accuracy = \dfrac{TP+TN}{TP+TN+FP+FN},
\end{aligned}
\end{equation}
TP is the number of pixels were segmented by the method and they are matching with the ground truth segmented region. FP is the number of pixels were segmented by the method and they are not matching with the ground truth segmented region. TN is the number of pixels were not segmented by the method and they should not be part of segmentation. FN is the number of pixels were not segmented by the method and they should be part of segmentation. The results of SLIC superpixel method following by the merging function are illustrated in Figures \ref{fig6} and \ref{fig7}. Figure \ref{fig8} shows the temporal segmentation accuracy for 50 consecutive frames for SLIC method with 1500 superpixels. \\
QS {\cite{vedaldi2008quick}} and Gb {\cite{Felzenszwalb04}} methods were selected to compare the SLIC with the common superpixel methods. The methods were examined using python platform on a MacBook pro 2.7 GHz Intel Core i5 with 8 GB 1867 MHz DDR3. Figure \ref{fig9} shows the results for sensitivity and the average speed of these three methods to segment the superpixels in a frame. \\
In addition, we used t-SNE to visualize a 2D representation of the extracted features mentioned in section \ref{sect:fet}. The results are illustrated in Fig \ref{fig10}. Finding the best features can help to design better trackers and this leads the goal needed in biomechanics and neuroscience studies. The t-SNE shows the automatic classification of the three groups can be easier for mice compared to rats, especially in differentiation between the body and the other regions. \\
We presented a simple tracker in section \ref{sect:tra} to show how the segmented regions can be used to design a tracker. We have found that this tracker can be used to track any of the objects but not paws. We evaluated the performance of this tracker on 5 trials from mice each having 1000 frames to track the lowest part of the ear. Out of 5000 frames, there was just 43 consecutive mistakes which happened when mouse was turning the head in one of the trials. 
\section{Discussion}
\label{sect:con}
We presented a method for segmentation of different parts of rodents body running on treadmill. We categorized the body parts to three classes: paw; ear, nose, and tail; and skin. the SLIC superpixels method was used for the segmentation and it was applied on three different color images (RGB, hue, and gray scale) from three different color spaces (RGB, HSV, and gray scale) with three SLIC sizes (500, 1500, and 4500). After segmentation, we calculated the average of intensity for each of the segments in the three images, and then, we connected superpixel segments to each other if they were neighbor and they had less than ten percent difference in average intensity. This process is illustrated in Figure \ref{fig3}.\\
By increasing the SLIC size, the accuracy of segmentation increased, especially for smaller objects; however, it costs the required time for the processes. This is shown in Figures \ref{fig1} and \ref{fig2}. \\
\begin{figure*}[t!p]
         \centering
      \includegraphics[scale=0.8]{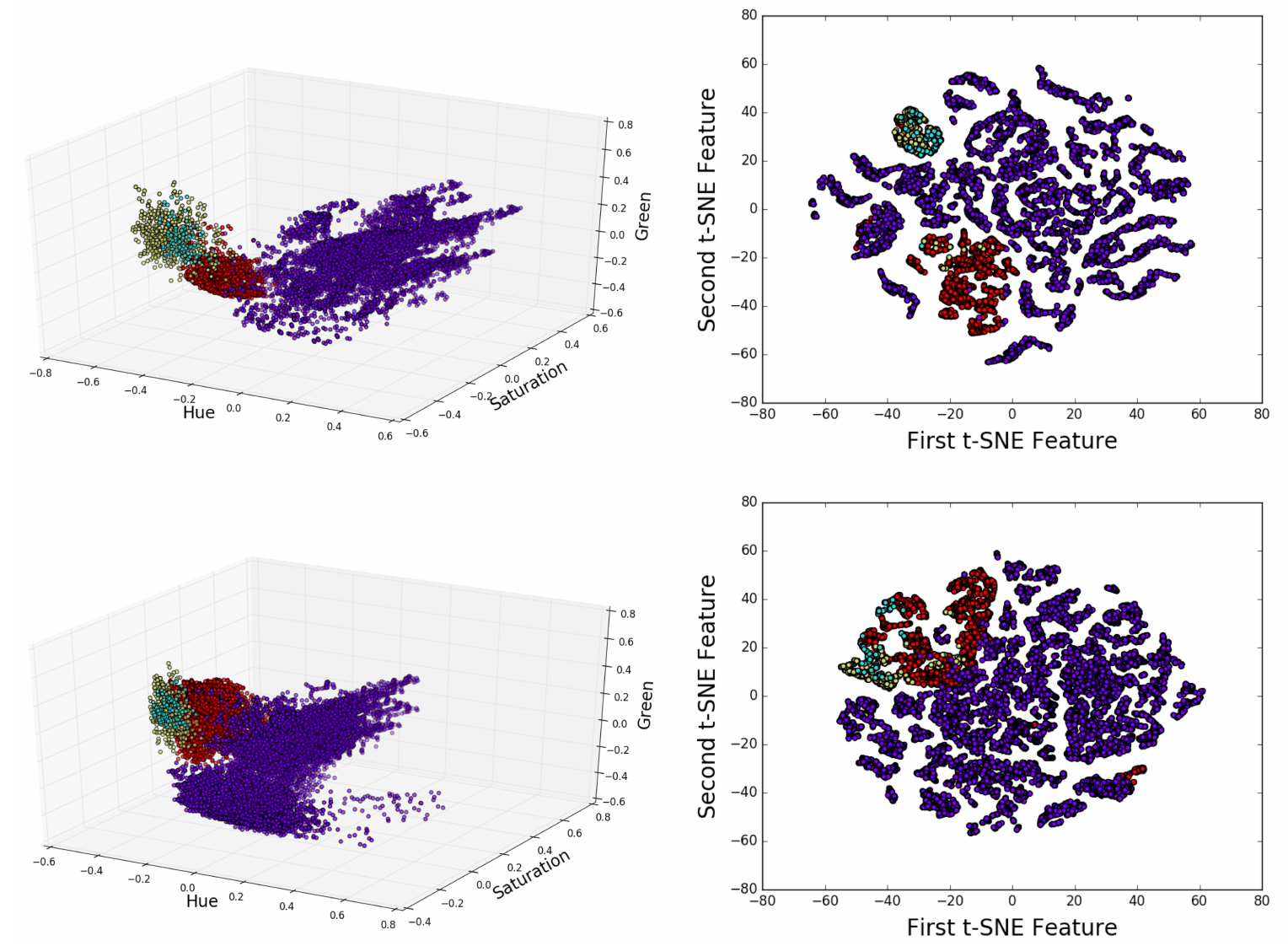}
\caption{The t-SNE graph. (a) and (c) show the scatter plots of three features (green channel, hue channel, and saturation channel averages) extracted from superpixels. (a) and (c) are respectively the results for 20 mice frames and 20 rats frames. (b) and (d) show respectively the t-SNE plot for 20 mice frames and 20 rats frames to demonstrates how the combination texture and color features can be used to classify the objects for the future studies. The purple, red, yellow, and light blue are respectively the background, body, three landmarks, and paws.}  \label{fig10}
\end{figure*}
Among the three color spaces selected, RGB showed the best accuracy of segmentation, although hue had almost the same results. This was more distinctive especially for lower SLIC sizes, as can be seen in Figures \ref{fig1} and \ref{fig2}. Therefore, the best image format for using SLIC in our context is RGB. \\
As mentioned above, we used the function, illustrated in Figure \ref{fig3}, to connect the segments to each other. This function gave us the possibility to join the segments with a similar range of average intensity values. We divided animals parts to three classes just to differentiate between these parts. Using this function, we created larger segments, and finally, the segments consisting three classes were automatically selected. The results are shown in Figures \ref{fig4} and \ref{fig5}. \\
Having these larger segments allowed us to compare the segmentation sensitivity, specificity, accuracy, and precision compared to the manually outlined for each frame. The results are illustrated in Figure \ref{fig6}. The results indicated that the sensitivity and precision of segmentation increased by having a larger number of superpixels. This trend was seen for the specificity and accuracy; however, they had smaller changes comparing to the other two measures because of the number of pixels indicating TN was larger compared to the other three variables (TP, FP, and FN), especially for mice. The changes for specificity and accuracy were more significant for rats because of the animal size, as seen in Figure \ref{fig7}. \\
As shown in Figure \ref{fig6}, The best image to segment body in mice was the RGB image while the best image for the segmentation of paw and tail was the hue channel. This pattern was not seen for rats. The best image was always the hue channel from the HSV color space, based on the reported results in Figure \ref{fig7}. In addition, to segment the body of rats, the gray scale image showed the higher measures compared to the RGB image; demonstrating that the fact that the white body of rats was easier to distinguish from the background.\\
In conclusion, the SLIC supper pixel gave reliable results for the segmentation of landmarks in rodents body running on the treadmill. RGB and HSV color spaces achieved almost similar segmented regions, although RGB was slightly better in the term of segmentation, especially for lower SLIC size numbers. This means that when we had bigger superpixels, creating more meaningful superpixels, the RGB images showed higher measures as can be seen in Figures \ref{fig6} and \ref{fig7}. This was opposite when it came to using color channels information for classifying the segmented region using the average intensity. Hue carried more information by itself compared to the average of R, G, and the gray scale. It gives us the idea to use RGB for segmentation and use hue channel information for classification in future works. \\
The results of tail segmentation (Figures \ref{fig6} and \ref{fig7}) showed a zig zag behavior in the ROC plots (especially the frames in the RGB color space) captured from both rats and mice (more significance changes for mice). This might be because the tail was small and narrow for some parts and differentiation of these small parts from background was harder using the average of RGB channels or gray scale intensity. In addition, the lateral part of tail showed a different color information compared with other parts (as shown Figure \ref{fig1}). \\
\begin{figure}[t!p]
         \centering
      \includegraphics[scale=0.7]{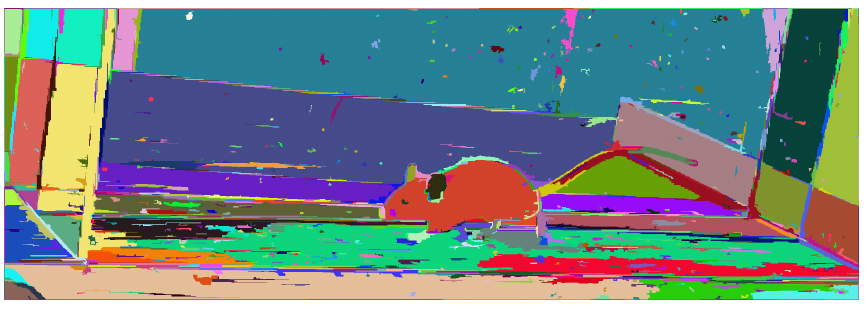}
\caption{A sample frame showing how the Gb method can segment the body and other regions without using the merging function. However, because of the relatively large image size, it takes more than 50 seconds to give the outcome.}  \label{fig11}
\end{figure}
As mentioned, there have been methods proposed to segment the animal using simple thresholding, cross correlation, or template matching {\cite{hedrick2008software, noldus2002computerised, noldus2001ethovision}}. These methods can provide information for behavioral experiments while tracking of specific landmarks on body is needed for biomechanics. The proposed method using SLIC provides remarkably fast and accurate segmentation leading to a promising tracking system as an example presented here.\\
On the other hand, superpixel based methods have been used frequently for detection of human hand and the gestures. The t-SNE was used to evaluate the importance of features for superpixels for hand detection {\cite{li2013pixel}}. This was inspired us to extract features and evaluate how much they can provide information to distinguish the regions from each other and background. The results are illustrated in Figure \ref{fig10}. \\
Last but not least, although SLIC was equally good for segmentation and much faster than the other algorithms. However, Gb can be used to segment the ear, paws, and body in a merged form by itself as seen in Figure \ref{fig11}. However, it takes more time to have the regions segmented and accuracy is lower comparing to SLIC. \\
For future directions, we will extract more texture, color, and kinematics features, and then, classify and track these regions using NN {\cite{maghsoudi2014informative}}, SVM {\cite{Maghsoudi16}}, or neuro fuzzy logic. Achieving this goal will help us to track each of the objects in the video, that subsequently will lead to an accurate 3D reconstruction of these objects. 3D data on animal movement will likely provide a wealth of information for not just biomechanics but also neuroscience and broader biological investigations. Finally, we will try to predic and edit the gait transitions {\cite{wilshin2017morphology}}.
\section*{Acknowledgments}
This material is based upon work supported by, or in part by, the U. S. Army Research Laboratory and the U. S. Army Research Office under contract/grant number W911NF1410141, proposal 64929EG, to A. Spence. 
\newpage
\bibliography{report}

\begin{thebibliography}{46}
\expandafter\ifx\csname natexlab\endcsname\relax\def\natexlab#1{#1}\fi
\expandafter\ifx\csname bibnamefont\endcsname\relax
  \def\bibnamefont#1{#1}\fi
\expandafter\ifx\csname bibfnamefont\endcsname\relax
  \def\bibfnamefont#1{#1}\fi
\expandafter\ifx\csname citenamefont\endcsname\relax
  \def\citenamefont#1{#1}\fi
\expandafter\ifx\csname url\endcsname\relax
  \def\url#1{\texttt{#1}}\fi
\expandafter\ifx\csname urlprefix\endcsname\relax\def\urlprefix{URL }\fi
\providecommand{\bibinfo}[2]{#2}
\providecommand{\eprint}[2][]{\url{#2}}

\bibitem[{\citenamefont{Arnold et~al.}(1992)\citenamefont{Arnold, McVey,
  Farrell, Deurloo, and Grasso}}]{Arnold92}
\bibinfo{author}{\bibfnamefont{P.~B.} \bibnamefont{Arnold}},
  \bibinfo{author}{\bibfnamefont{P.~P.} \bibnamefont{McVey}},
  \bibinfo{author}{\bibfnamefont{W.~J.} \bibnamefont{Farrell}},
  \bibinfo{author}{\bibfnamefont{T.~M.} \bibnamefont{Deurloo}},
  \bibnamefont{and} \bibinfo{author}{\bibfnamefont{A.~R.}
  \bibnamefont{Grasso}}, \bibinfo{journal}{Archives of physical medicine and
  rehabilitation} \textbf{\bibinfo{volume}{73}}, \bibinfo{pages}{665}
  (\bibinfo{year}{1992}).

\bibitem[{\citenamefont{Cirak et~al.}(2011)\citenamefont{Cirak,
  Arechavala-Gomeza, Guglieri, Feng, Torelli, Anthony, Abbs, Garralda, Bourke,
  Wells et~al.}}]{Cirak11}
\bibinfo{author}{\bibfnamefont{S.}~\bibnamefont{Cirak}},
  \bibinfo{author}{\bibfnamefont{V.}~\bibnamefont{Arechavala-Gomeza}},
  \bibinfo{author}{\bibfnamefont{M.}~\bibnamefont{Guglieri}},
  \bibinfo{author}{\bibfnamefont{L.}~\bibnamefont{Feng}},
  \bibinfo{author}{\bibfnamefont{S.}~\bibnamefont{Torelli}},
  \bibinfo{author}{\bibfnamefont{K.}~\bibnamefont{Anthony}},
  \bibinfo{author}{\bibfnamefont{S.}~\bibnamefont{Abbs}},
  \bibinfo{author}{\bibfnamefont{M.~E.} \bibnamefont{Garralda}},
  \bibinfo{author}{\bibfnamefont{J.}~\bibnamefont{Bourke}},
  \bibinfo{author}{\bibfnamefont{D.~J.} \bibnamefont{Wells}},
  \bibnamefont{et~al.}, \bibinfo{journal}{The Lancet}
  \textbf{\bibinfo{volume}{378}}, \bibinfo{pages}{595} (\bibinfo{year}{2011}).

\bibitem[{\citenamefont{Herr and Grabowski}(2012)}]{Herr12}
\bibinfo{author}{\bibfnamefont{H.~M.} \bibnamefont{Herr}} \bibnamefont{and}
  \bibinfo{author}{\bibfnamefont{A.~M.} \bibnamefont{Grabowski}}, in
  \emph{\bibinfo{booktitle}{Proc. R. Soc. B}} (\bibinfo{organization}{The Royal
  Society}, \bibinfo{year}{2012}), vol. \bibinfo{volume}{279}, pp.
  \bibinfo{pages}{457--464}.

\bibitem[{\citenamefont{RAIBERT}(2012)}]{Alphadog12}
\bibinfo{author}{\bibfnamefont{M.}~\bibnamefont{RAIBERT}}, in
  \emph{\bibinfo{booktitle}{Adaptive Mobile Robotics: Proceedings of the 15th
  International Conference on Climbing and Walking Robots and the Support
  Technologies for Mobile Machines, Baltimore, USA, 23-26 July, 2012}}
  (\bibinfo{organization}{World Scientific}, \bibinfo{year}{2012}),
  p.~\bibinfo{pages}{7}.

\bibitem[{\citenamefont{Clarke and Still}(1999)}]{Clarke99}
\bibinfo{author}{\bibfnamefont{K.}~\bibnamefont{Clarke}} \bibnamefont{and}
  \bibinfo{author}{\bibfnamefont{J.}~\bibnamefont{Still}},
  \bibinfo{journal}{Physiology \& Behavior} \textbf{\bibinfo{volume}{66}},
  \bibinfo{pages}{723} (\bibinfo{year}{1999}).

\bibitem[{\citenamefont{Orlovski et~al.}(1999)\citenamefont{Orlovski,
  Deliagina, and Grillner}}]{Orlovsky99}
\bibinfo{author}{\bibfnamefont{G.~N.} \bibnamefont{Orlovski}},
  \bibinfo{author}{\bibfnamefont{T.}~\bibnamefont{Deliagina}},
  \bibnamefont{and} \bibinfo{author}{\bibfnamefont{S.}~\bibnamefont{Grillner}},
  \emph{\bibinfo{title}{Neuronal control of locomotion: from mollusc to man}}
  (\bibinfo{publisher}{Oxford University Press}, \bibinfo{year}{1999}).

\bibitem[{\citenamefont{Courtine et~al.}(2008)\citenamefont{Courtine, Song,
  Roy, Zhong, mann, Ao, Qi, Edgerton, and Sofroniew}}]{courtine2008recovery}
\bibinfo{author}{\bibfnamefont{G.}~\bibnamefont{Courtine}},
  \bibinfo{author}{\bibfnamefont{B.}~\bibnamefont{Song}},
  \bibinfo{author}{\bibfnamefont{R.~R.} \bibnamefont{Roy}},
  \bibinfo{author}{\bibfnamefont{H.}~\bibnamefont{Zhong}},
  \bibinfo{author}{\bibfnamefont{J.~E.} \bibnamefont{mann}},
  \bibinfo{author}{\bibfnamefont{Y.}~\bibnamefont{Ao}},
  \bibinfo{author}{\bibfnamefont{J.}~\bibnamefont{Qi}},
  \bibinfo{author}{\bibfnamefont{V.~R.} \bibnamefont{Edgerton}},
  \bibnamefont{and} \bibinfo{author}{\bibfnamefont{M.~V.}
  \bibnamefont{Sofroniew}}, \bibinfo{journal}{Nature medicine}
  \textbf{\bibinfo{volume}{14}}, \bibinfo{pages}{69} (\bibinfo{year}{2008}).

\bibitem[{\citenamefont{Migliaccio et~al.}(2011)\citenamefont{Migliaccio,
  Meierhofer, and Della~Santina}}]{migliaccio11}
\bibinfo{author}{\bibfnamefont{A.~A.} \bibnamefont{Migliaccio}},
  \bibinfo{author}{\bibfnamefont{R.}~\bibnamefont{Meierhofer}},
  \bibnamefont{and} \bibinfo{author}{\bibfnamefont{C.~C.}
  \bibnamefont{Della~Santina}}, \bibinfo{journal}{Experimental brain research}
  \textbf{\bibinfo{volume}{210}}, \bibinfo{pages}{489} (\bibinfo{year}{2011}).

\bibitem[{\citenamefont{Baker}(2005)}]{baker05}
\bibinfo{author}{\bibfnamefont{J.~F.} \bibnamefont{Baker}},
  \bibinfo{journal}{Experimental brain research}
  \textbf{\bibinfo{volume}{167}}, \bibinfo{pages}{108} (\bibinfo{year}{2005}).

\bibitem[{\citenamefont{Revzen and Guckenheimer}(2012)}]{Revzen12}
\bibinfo{author}{\bibfnamefont{S.}~\bibnamefont{Revzen}} \bibnamefont{and}
  \bibinfo{author}{\bibfnamefont{J.~M.} \bibnamefont{Guckenheimer}},
  \bibinfo{journal}{Journal of The Royal Society Interface}
  \textbf{\bibinfo{volume}{9}}, \bibinfo{pages}{957} (\bibinfo{year}{2012}).

\bibitem[{\citenamefont{Wiltschko et~al.}(2015)\citenamefont{Wiltschko,
  Johnson, Iurilli, Peterson, Katon, Pashkovski, Abraira, Adams, and
  Datta}}]{Wiltschko15}
\bibinfo{author}{\bibfnamefont{A.~B.} \bibnamefont{Wiltschko}},
  \bibinfo{author}{\bibfnamefont{M.~J.} \bibnamefont{Johnson}},
  \bibinfo{author}{\bibfnamefont{G.}~\bibnamefont{Iurilli}},
  \bibinfo{author}{\bibfnamefont{R.~E.} \bibnamefont{Peterson}},
  \bibinfo{author}{\bibfnamefont{J.~M.} \bibnamefont{Katon}},
  \bibinfo{author}{\bibfnamefont{S.~L.} \bibnamefont{Pashkovski}},
  \bibinfo{author}{\bibfnamefont{V.~E.} \bibnamefont{Abraira}},
  \bibinfo{author}{\bibfnamefont{R.~P.} \bibnamefont{Adams}}, \bibnamefont{and}
  \bibinfo{author}{\bibfnamefont{S.~R.} \bibnamefont{Datta}},
  \bibinfo{journal}{Neuron} \textbf{\bibinfo{volume}{88}},
  \bibinfo{pages}{1121} (\bibinfo{year}{2015}).

\bibitem[{\citenamefont{Hedrick}(2008)}]{hedrick2008software}
\bibinfo{author}{\bibfnamefont{T.~L.} \bibnamefont{Hedrick}},
  \bibinfo{journal}{Bioinspiration \& biomimetics}
  \textbf{\bibinfo{volume}{3}}, \bibinfo{pages}{034001} (\bibinfo{year}{2008}).

\bibitem[{\citenamefont{Noldus et~al.}(2002)\citenamefont{Noldus, Spink, and
  Tegelenbosch}}]{noldus2002computerised}
\bibinfo{author}{\bibfnamefont{L.~P.} \bibnamefont{Noldus}},
  \bibinfo{author}{\bibfnamefont{A.~J.} \bibnamefont{Spink}}, \bibnamefont{and}
  \bibinfo{author}{\bibfnamefont{R.~A.} \bibnamefont{Tegelenbosch}},
  \bibinfo{journal}{Computers and Electronics in Agriculture}
  \textbf{\bibinfo{volume}{35}}, \bibinfo{pages}{201} (\bibinfo{year}{2002}).

\bibitem[{\citenamefont{Wenger et~al.}(2016)\citenamefont{Wenger, Moraud,
  Gandar, Musienko, Capogrosso, Baud, Le~Goff, Barraud, Pavlova, Dominici
  et~al.}}]{wenger2016spatiotemporal}
\bibinfo{author}{\bibfnamefont{N.}~\bibnamefont{Wenger}},
  \bibinfo{author}{\bibfnamefont{E.~M.} \bibnamefont{Moraud}},
  \bibinfo{author}{\bibfnamefont{J.}~\bibnamefont{Gandar}},
  \bibinfo{author}{\bibfnamefont{P.}~\bibnamefont{Musienko}},
  \bibinfo{author}{\bibfnamefont{M.}~\bibnamefont{Capogrosso}},
  \bibinfo{author}{\bibfnamefont{L.}~\bibnamefont{Baud}},
  \bibinfo{author}{\bibfnamefont{C.~G.} \bibnamefont{Le~Goff}},
  \bibinfo{author}{\bibfnamefont{Q.}~\bibnamefont{Barraud}},
  \bibinfo{author}{\bibfnamefont{N.}~\bibnamefont{Pavlova}},
  \bibinfo{author}{\bibfnamefont{N.}~\bibnamefont{Dominici}},
  \bibnamefont{et~al.}, \bibinfo{journal}{Nature medicine}
  \textbf{\bibinfo{volume}{22}}, \bibinfo{pages}{138} (\bibinfo{year}{2016}).

\bibitem[{\citenamefont{Dorman et~al.}(2014)\citenamefont{Dorman, Krug,
  Frizelle, Funkenbusch, and Mahowald}}]{Dorman14}
\bibinfo{author}{\bibfnamefont{C.~W.} \bibnamefont{Dorman}},
  \bibinfo{author}{\bibfnamefont{H.~E.} \bibnamefont{Krug}},
  \bibinfo{author}{\bibfnamefont{S.~P.} \bibnamefont{Frizelle}},
  \bibinfo{author}{\bibfnamefont{S.}~\bibnamefont{Funkenbusch}},
  \bibnamefont{and} \bibinfo{author}{\bibfnamefont{M.~L.}
  \bibnamefont{Mahowald}}, \bibinfo{journal}{Journal of pain research}
  \textbf{\bibinfo{volume}{7}}, \bibinfo{pages}{25} (\bibinfo{year}{2014}).

\bibitem[{\citenamefont{Gadalla et~al.}(2014)\citenamefont{Gadalla, Ross,
  Riddell, Bailey, and Cobb}}]{Gadalla14}
\bibinfo{author}{\bibfnamefont{K.~K.} \bibnamefont{Gadalla}},
  \bibinfo{author}{\bibfnamefont{P.~D.} \bibnamefont{Ross}},
  \bibinfo{author}{\bibfnamefont{J.~S.} \bibnamefont{Riddell}},
  \bibinfo{author}{\bibfnamefont{M.~E.} \bibnamefont{Bailey}},
  \bibnamefont{and} \bibinfo{author}{\bibfnamefont{S.~R.} \bibnamefont{Cobb}},
  \bibinfo{journal}{PloS one} \textbf{\bibinfo{volume}{9}},
  \bibinfo{pages}{e112889} (\bibinfo{year}{2014}).

\bibitem[{\citenamefont{Huehnchen et~al.}(2013)\citenamefont{Huehnchen,
  Boehmerle, and Endres}}]{Huehnchen13}
\bibinfo{author}{\bibfnamefont{P.}~\bibnamefont{Huehnchen}},
  \bibinfo{author}{\bibfnamefont{W.}~\bibnamefont{Boehmerle}},
  \bibnamefont{and} \bibinfo{author}{\bibfnamefont{M.}~\bibnamefont{Endres}},
  \bibinfo{journal}{PloS one} \textbf{\bibinfo{volume}{8}},
  \bibinfo{pages}{e76772} (\bibinfo{year}{2013}).

\bibitem[{\citenamefont{Hamers et~al.}(2001)\citenamefont{Hamers, Lankhorst,
  van Laar, Veldhuis, and Gispen}}]{Hamers04}
\bibinfo{author}{\bibfnamefont{F.~P.} \bibnamefont{Hamers}},
  \bibinfo{author}{\bibfnamefont{A.~J.} \bibnamefont{Lankhorst}},
  \bibinfo{author}{\bibfnamefont{T.~J.} \bibnamefont{van Laar}},
  \bibinfo{author}{\bibfnamefont{W.~B.} \bibnamefont{Veldhuis}},
  \bibnamefont{and} \bibinfo{author}{\bibfnamefont{W.~H.}
  \bibnamefont{Gispen}}, \bibinfo{journal}{Journal of neurotrauma}
  \textbf{\bibinfo{volume}{18}}, \bibinfo{pages}{187} (\bibinfo{year}{2001}).

\bibitem[{\citenamefont{Parvathy and Masocha}(2013)}]{Parvathy13}
\bibinfo{author}{\bibfnamefont{S.~S.} \bibnamefont{Parvathy}} \bibnamefont{and}
  \bibinfo{author}{\bibfnamefont{W.}~\bibnamefont{Masocha}},
  \bibinfo{journal}{BMC musculoskeletal disorders}
  \textbf{\bibinfo{volume}{14}}, \bibinfo{pages}{14} (\bibinfo{year}{2013}).

\bibitem[{\citenamefont{Maghsoudi et~al.}(2015)\citenamefont{Maghsoudi,
  Tabrizi, Robertson, Shamble, and Spence}}]{Maghsoudi15}
\bibinfo{author}{\bibfnamefont{O.~H.} \bibnamefont{Maghsoudi}},
  \bibinfo{author}{\bibfnamefont{A.~V.} \bibnamefont{Tabrizi}},
  \bibinfo{author}{\bibfnamefont{B.}~\bibnamefont{Robertson}},
  \bibinfo{author}{\bibfnamefont{P.}~\bibnamefont{Shamble}}, \bibnamefont{and}
  \bibinfo{author}{\bibfnamefont{A.}~\bibnamefont{Spence}}, in
  \emph{\bibinfo{booktitle}{Signal Processing in Medicine and Biology Symposium
  (SPMB), 2015 IEEE}} (\bibinfo{organization}{IEEE}, \bibinfo{year}{2015}), pp.
  \bibinfo{pages}{1--2}.

\bibitem[{\citenamefont{Spence et~al.}(2013)\citenamefont{Spence,
  Nicholson-Thomas, and Lampe}}]{Spence13}
\bibinfo{author}{\bibfnamefont{A.~J.} \bibnamefont{Spence}},
  \bibinfo{author}{\bibfnamefont{G.}~\bibnamefont{Nicholson-Thomas}},
  \bibnamefont{and} \bibinfo{author}{\bibfnamefont{R.}~\bibnamefont{Lampe}},
  \bibinfo{journal}{Journal of neuroscience methods}
  \textbf{\bibinfo{volume}{215}}, \bibinfo{pages}{164} (\bibinfo{year}{2013}).

\bibitem[{\citenamefont{Gon{\c{c}}alves
  et~al.}(2007)\citenamefont{Gon{\c{c}}alves, Monteiro, de~Andrade~Silva,
  Machado, Pistori, and Odakura}}]{gonccalves07}
\bibinfo{author}{\bibfnamefont{W.~N.} \bibnamefont{Gon{\c{c}}alves}},
  \bibinfo{author}{\bibfnamefont{J.~B.~O.} \bibnamefont{Monteiro}},
  \bibinfo{author}{\bibfnamefont{J.}~\bibnamefont{de~Andrade~Silva}},
  \bibinfo{author}{\bibfnamefont{B.~B.} \bibnamefont{Machado}},
  \bibinfo{author}{\bibfnamefont{H.}~\bibnamefont{Pistori}}, \bibnamefont{and}
  \bibinfo{author}{\bibfnamefont{V.}~\bibnamefont{Odakura}}, in
  \emph{\bibinfo{booktitle}{Computer Graphics and Image Processing, 2007.
  SIBGRAPI 2007. XX Brazilian Symposium on}} (\bibinfo{organization}{IEEE},
  \bibinfo{year}{2007}), pp. \bibinfo{pages}{173--178}.

\bibitem[{\citenamefont{Pistori et~al.}(2010)\citenamefont{Pistori, Odakura,
  Monteiro, Gon{\c{c}}alves, Roel, de~Andrade~Silva, and Machado}}]{pistori10}
\bibinfo{author}{\bibfnamefont{H.}~\bibnamefont{Pistori}},
  \bibinfo{author}{\bibfnamefont{V.~V. V.~A.} \bibnamefont{Odakura}},
  \bibinfo{author}{\bibfnamefont{J.~B.~O.} \bibnamefont{Monteiro}},
  \bibinfo{author}{\bibfnamefont{W.~N.} \bibnamefont{Gon{\c{c}}alves}},
  \bibinfo{author}{\bibfnamefont{A.~R.} \bibnamefont{Roel}},
  \bibinfo{author}{\bibfnamefont{J.}~\bibnamefont{de~Andrade~Silva}},
  \bibnamefont{and} \bibinfo{author}{\bibfnamefont{B.~B.}
  \bibnamefont{Machado}}, \bibinfo{journal}{Pattern Recognition Letters}
  \textbf{\bibinfo{volume}{31}}, \bibinfo{pages}{337} (\bibinfo{year}{2010}).

\bibitem[{\citenamefont{Ren and Malik}(2003)}]{Ren03}
\bibinfo{author}{\bibfnamefont{X.}~\bibnamefont{Ren}} \bibnamefont{and}
  \bibinfo{author}{\bibfnamefont{J.}~\bibnamefont{Malik}}, in
  \emph{\bibinfo{booktitle}{ICCV}} (\bibinfo{year}{2003}),
  vol.~\bibinfo{volume}{1}, pp. \bibinfo{pages}{10--17}.

\bibitem[{\citenamefont{Comaniciu and Meer}(2002)}]{Comaniciu02}
\bibinfo{author}{\bibfnamefont{D.}~\bibnamefont{Comaniciu}} \bibnamefont{and}
  \bibinfo{author}{\bibfnamefont{P.}~\bibnamefont{Meer}},
  \bibinfo{journal}{IEEE Transactions on pattern analysis and machine
  intelligence} \textbf{\bibinfo{volume}{24}}, \bibinfo{pages}{603}
  (\bibinfo{year}{2002}).

\bibitem[{\citenamefont{Felzenszwalb and Huttenlocher}(2004)}]{Felzenszwalb04}
\bibinfo{author}{\bibfnamefont{P.~F.} \bibnamefont{Felzenszwalb}}
  \bibnamefont{and} \bibinfo{author}{\bibfnamefont{D.~P.}
  \bibnamefont{Huttenlocher}}, \bibinfo{journal}{International journal of
  computer vision} \textbf{\bibinfo{volume}{59}}, \bibinfo{pages}{167}
  (\bibinfo{year}{2004}).

\bibitem[{\citenamefont{Levinshtein et~al.}(2009)\citenamefont{Levinshtein,
  Stere, Kutulakos, Fleet, Dickinson, and Siddiqi}}]{Levinshtein09}
\bibinfo{author}{\bibfnamefont{A.}~\bibnamefont{Levinshtein}},
  \bibinfo{author}{\bibfnamefont{A.}~\bibnamefont{Stere}},
  \bibinfo{author}{\bibfnamefont{K.~N.} \bibnamefont{Kutulakos}},
  \bibinfo{author}{\bibfnamefont{D.~J.} \bibnamefont{Fleet}},
  \bibinfo{author}{\bibfnamefont{S.~J.} \bibnamefont{Dickinson}},
  \bibnamefont{and} \bibinfo{author}{\bibfnamefont{K.}~\bibnamefont{Siddiqi}},
  \bibinfo{journal}{IEEE transactions on pattern analysis and machine
  intelligence} \textbf{\bibinfo{volume}{31}}, \bibinfo{pages}{2290}
  (\bibinfo{year}{2009}).

\bibitem[{\citenamefont{Achanta et~al.}(2012)\citenamefont{Achanta, Shaji,
  Smith, Lucchi, Fua, and S{\"u}sstrunk}}]{Achanta12}
\bibinfo{author}{\bibfnamefont{R.}~\bibnamefont{Achanta}},
  \bibinfo{author}{\bibfnamefont{A.}~\bibnamefont{Shaji}},
  \bibinfo{author}{\bibfnamefont{K.}~\bibnamefont{Smith}},
  \bibinfo{author}{\bibfnamefont{A.}~\bibnamefont{Lucchi}},
  \bibinfo{author}{\bibfnamefont{P.}~\bibnamefont{Fua}}, \bibnamefont{and}
  \bibinfo{author}{\bibfnamefont{S.}~\bibnamefont{S{\"u}sstrunk}},
  \bibinfo{journal}{IEEE transactions on pattern analysis and machine
  intelligence} \textbf{\bibinfo{volume}{34}}, \bibinfo{pages}{2274}
  (\bibinfo{year}{2012}).

\bibitem[{\citenamefont{Veksler et~al.}(2010)\citenamefont{Veksler, Boykov, and
  Mehrani}}]{Veksler10}
\bibinfo{author}{\bibfnamefont{O.}~\bibnamefont{Veksler}},
  \bibinfo{author}{\bibfnamefont{Y.}~\bibnamefont{Boykov}}, \bibnamefont{and}
  \bibinfo{author}{\bibfnamefont{P.}~\bibnamefont{Mehrani}}, in
  \emph{\bibinfo{booktitle}{European conference on Computer vision}}
  (\bibinfo{organization}{Springer}, \bibinfo{year}{2010}), pp.
  \bibinfo{pages}{211--224}.

\bibitem[{\citenamefont{Shu et~al.}(2013)\citenamefont{Shu, Dehghan, and
  Shah}}]{shu2013improving}
\bibinfo{author}{\bibfnamefont{G.}~\bibnamefont{Shu}},
  \bibinfo{author}{\bibfnamefont{A.}~\bibnamefont{Dehghan}}, \bibnamefont{and}
  \bibinfo{author}{\bibfnamefont{M.}~\bibnamefont{Shah}}, in
  \emph{\bibinfo{booktitle}{Proceedings of the IEEE Conference on Computer
  Vision and Pattern Recognition}} (\bibinfo{year}{2013}), pp.
  \bibinfo{pages}{3721--3727}.

\bibitem[{\citenamefont{Serra et~al.}(2013)\citenamefont{Serra, Camurri,
  Baraldi, Benedetti, and Cucchiara}}]{serra2013hand}
\bibinfo{author}{\bibfnamefont{G.}~\bibnamefont{Serra}},
  \bibinfo{author}{\bibfnamefont{M.}~\bibnamefont{Camurri}},
  \bibinfo{author}{\bibfnamefont{L.}~\bibnamefont{Baraldi}},
  \bibinfo{author}{\bibfnamefont{M.}~\bibnamefont{Benedetti}},
  \bibnamefont{and}
  \bibinfo{author}{\bibfnamefont{R.}~\bibnamefont{Cucchiara}}, in
  \emph{\bibinfo{booktitle}{Proceedings of the 3rd ACM international workshop
  on Interactive multimedia on mobile \& portable devices}}
  (\bibinfo{organization}{ACM}, \bibinfo{year}{2013}), pp.
  \bibinfo{pages}{31--36}.

\bibitem[{\citenamefont{Li and Kitani}(2013{\natexlab{a}})}]{li2013model}
\bibinfo{author}{\bibfnamefont{C.}~\bibnamefont{Li}} \bibnamefont{and}
  \bibinfo{author}{\bibfnamefont{K.~M.} \bibnamefont{Kitani}}, in
  \emph{\bibinfo{booktitle}{Proceedings of the IEEE International Conference on
  Computer Vision}} (\bibinfo{year}{2013}{\natexlab{a}}), pp.
  \bibinfo{pages}{2624--2631}.

\bibitem[{\citenamefont{Baraldi et~al.}(2014)\citenamefont{Baraldi, Paci,
  Serra, Benini, and Cucchiara}}]{baraldi2014gesture}
\bibinfo{author}{\bibfnamefont{L.}~\bibnamefont{Baraldi}},
  \bibinfo{author}{\bibfnamefont{F.}~\bibnamefont{Paci}},
  \bibinfo{author}{\bibfnamefont{G.}~\bibnamefont{Serra}},
  \bibinfo{author}{\bibfnamefont{L.}~\bibnamefont{Benini}}, \bibnamefont{and}
  \bibinfo{author}{\bibfnamefont{R.}~\bibnamefont{Cucchiara}}, in
  \emph{\bibinfo{booktitle}{Proceedings of the IEEE Conference on Computer
  Vision and Pattern Recognition Workshops}} (\bibinfo{year}{2014}), pp.
  \bibinfo{pages}{688--693}.

\bibitem[{\citenamefont{Li and Kitani}(2013{\natexlab{b}})}]{li2013pixel}
\bibinfo{author}{\bibfnamefont{C.}~\bibnamefont{Li}} \bibnamefont{and}
  \bibinfo{author}{\bibfnamefont{K.~M.} \bibnamefont{Kitani}}, in
  \emph{\bibinfo{booktitle}{Proceedings of the IEEE Conference on Computer
  Vision and Pattern Recognition}} (\bibinfo{year}{2013}{\natexlab{b}}), pp.
  \bibinfo{pages}{3570--3577}.

\bibitem[{\citenamefont{Noldus et~al.}(2001)\citenamefont{Noldus, Spink, and
  Tegelenbosch}}]{noldus2001ethovision}
\bibinfo{author}{\bibfnamefont{L.~P.} \bibnamefont{Noldus}},
  \bibinfo{author}{\bibfnamefont{A.~J.} \bibnamefont{Spink}}, \bibnamefont{and}
  \bibinfo{author}{\bibfnamefont{R.~A.} \bibnamefont{Tegelenbosch}},
  \bibinfo{journal}{Behavior Research Methods} \textbf{\bibinfo{volume}{33}},
  \bibinfo{pages}{398} (\bibinfo{year}{2001}).

\bibitem[{\citenamefont{Vedaldi and Soatto}(2008)}]{vedaldi2008quick}
\bibinfo{author}{\bibfnamefont{A.}~\bibnamefont{Vedaldi}} \bibnamefont{and}
  \bibinfo{author}{\bibfnamefont{S.}~\bibnamefont{Soatto}},
  \bibinfo{journal}{Computer vision--ECCV 2008} pp. \bibinfo{pages}{705--718}
  (\bibinfo{year}{2008}).

\bibitem[{\citenamefont{Maghsoudi
  et~al.}(2016{\natexlab{a}})\citenamefont{Maghsoudi, Alizadeh, and
  Mirmomen}}]{Maghsoudi16_2}
\bibinfo{author}{\bibfnamefont{O.~H.} \bibnamefont{Maghsoudi}},
  \bibinfo{author}{\bibfnamefont{M.}~\bibnamefont{Alizadeh}}, \bibnamefont{and}
  \bibinfo{author}{\bibfnamefont{M.}~\bibnamefont{Mirmomen}}, in
  \emph{\bibinfo{booktitle}{Signal Processing in Medicine and Biology Symposium
  (SPMB), 2016 IEEE}} (\bibinfo{organization}{IEEE},
  \bibinfo{year}{2016}{\natexlab{a}}), pp. \bibinfo{pages}{1--6}.

\bibitem[{\citenamefont{Straw et~al.}(2010)\citenamefont{Straw, Branson,
  Neumann, and Dickinson}}]{Straw11}
\bibinfo{author}{\bibfnamefont{A.~D.} \bibnamefont{Straw}},
  \bibinfo{author}{\bibfnamefont{K.}~\bibnamefont{Branson}},
  \bibinfo{author}{\bibfnamefont{T.~R.} \bibnamefont{Neumann}},
  \bibnamefont{and} \bibinfo{author}{\bibfnamefont{M.~H.}
  \bibnamefont{Dickinson}}, \bibinfo{journal}{Journal of The Royal Society
  Interface} p. \bibinfo{pages}{rsif20100230} (\bibinfo{year}{2010}).

\bibitem[{\citenamefont{Straw}(2017)}]{StrawGit}
\bibinfo{author}{\bibfnamefont{A.~D.} \bibnamefont{Straw}},
  \emph{\bibinfo{title}{Triggerbox}} (\bibinfo{year}{2017}),
  \urlprefix\url{https://github.com/strawlab/triggerbox}.

\bibitem[{\citenamefont{Maghsoudi
  et~al.}(2016{\natexlab{b}})\citenamefont{Maghsoudi, Tabrizi, Robertson,
  Shamble, and Spence}}]{Maghsoudi16}
\bibinfo{author}{\bibfnamefont{O.~H.} \bibnamefont{Maghsoudi}},
  \bibinfo{author}{\bibfnamefont{A.~V.} \bibnamefont{Tabrizi}},
  \bibinfo{author}{\bibfnamefont{B.}~\bibnamefont{Robertson}},
  \bibinfo{author}{\bibfnamefont{P.}~\bibnamefont{Shamble}}, \bibnamefont{and}
  \bibinfo{author}{\bibfnamefont{A.}~\bibnamefont{Spence}}, in
  \emph{\bibinfo{booktitle}{Signal Processing in Medicine and Biology Symposium
  (SPMB), 2016 IEEE}} (\bibinfo{organization}{IEEE},
  \bibinfo{year}{2016}{\natexlab{b}}), pp. \bibinfo{pages}{1--3}.

\bibitem[{\citenamefont{Mori et~al.}(2004)\citenamefont{Mori, Ren, Efros, and
  Malik}}]{Mori04}
\bibinfo{author}{\bibfnamefont{G.}~\bibnamefont{Mori}},
  \bibinfo{author}{\bibfnamefont{X.}~\bibnamefont{Ren}},
  \bibinfo{author}{\bibfnamefont{A.~A.} \bibnamefont{Efros}}, \bibnamefont{and}
  \bibinfo{author}{\bibfnamefont{J.}~\bibnamefont{Malik}}, in
  \emph{\bibinfo{booktitle}{Computer Vision and Pattern Recognition, 2004. CVPR
  2004. Proceedings of the 2004 IEEE Computer Society Conference on}}
  (\bibinfo{organization}{IEEE}, \bibinfo{year}{2004}),
  vol.~\bibinfo{volume}{2}, pp. \bibinfo{pages}{II--II}.

\bibitem[{\citenamefont{Li et~al.}(2012)\citenamefont{Li, Wu, and
  Chang}}]{Li12}
\bibinfo{author}{\bibfnamefont{Z.}~\bibnamefont{Li}},
  \bibinfo{author}{\bibfnamefont{X.-M.} \bibnamefont{Wu}}, \bibnamefont{and}
  \bibinfo{author}{\bibfnamefont{S.-F.} \bibnamefont{Chang}}, in
  \emph{\bibinfo{booktitle}{Computer Vision and Pattern Recognition (CVPR),
  2012 IEEE Conference on}} (\bibinfo{organization}{IEEE},
  \bibinfo{year}{2012}), pp. \bibinfo{pages}{789--796}.

\bibitem[{\citenamefont{Phung et~al.}(2005)\citenamefont{Phung, Bouzerdoum, and
  Chai}}]{Phung05}
\bibinfo{author}{\bibfnamefont{S.~L.} \bibnamefont{Phung}},
  \bibinfo{author}{\bibfnamefont{A.}~\bibnamefont{Bouzerdoum}},
  \bibnamefont{and} \bibinfo{author}{\bibfnamefont{D.}~\bibnamefont{Chai}},
  \bibinfo{journal}{IEEE transactions on pattern analysis and machine
  intelligence} \textbf{\bibinfo{volume}{27}}, \bibinfo{pages}{148}
  (\bibinfo{year}{2005}).

\bibitem[{\citenamefont{Albregtsen et~al.}(2008)}]{albregtsen2008statistical}
\bibinfo{author}{\bibfnamefont{F.}~\bibnamefont{Albregtsen}}
  \bibnamefont{et~al.}, \bibinfo{journal}{Image processing laboratory,
  department of informatics, university of oslo} \textbf{\bibinfo{volume}{5}}
  (\bibinfo{year}{2008}).

\bibitem[{\citenamefont{Maghsoudi et~al.}(2014)\citenamefont{Maghsoudi,
  Talebpour, Soltanian-Zadeh, Alizadeh, and
  Soleimani}}]{maghsoudi2014informative}
\bibinfo{author}{\bibfnamefont{O.~H.} \bibnamefont{Maghsoudi}},
  \bibinfo{author}{\bibfnamefont{A.}~\bibnamefont{Talebpour}},
  \bibinfo{author}{\bibfnamefont{H.}~\bibnamefont{Soltanian-Zadeh}},
  \bibinfo{author}{\bibfnamefont{M.}~\bibnamefont{Alizadeh}}, \bibnamefont{and}
  \bibinfo{author}{\bibfnamefont{H.~A.} \bibnamefont{Soleimani}},
  \bibinfo{journal}{Journal of Advanced Computing}
  \textbf{\bibinfo{volume}{3}}, \bibinfo{pages}{12} (\bibinfo{year}{2014}).

\bibitem[{\citenamefont{Wilshin et~al.}(2017)\citenamefont{Wilshin, Haynes,
  Porteous, Koditschek, Revzen, and Spence}}]{wilshin2017morphology}
\bibinfo{author}{\bibfnamefont{S.}~\bibnamefont{Wilshin}},
  \bibinfo{author}{\bibfnamefont{G.~C.} \bibnamefont{Haynes}},
  \bibinfo{author}{\bibfnamefont{J.}~\bibnamefont{Porteous}},
  \bibinfo{author}{\bibfnamefont{D.}~\bibnamefont{Koditschek}},
  \bibinfo{author}{\bibfnamefont{S.}~\bibnamefont{Revzen}}, \bibnamefont{and}
  \bibinfo{author}{\bibfnamefont{A.~J.} \bibnamefont{Spence}},
  \bibinfo{journal}{Biological cybernetics}  (\bibinfo{year}{2017}).

\end{thebibliography}

\section*{Biography}
\textbf{Omid Haji Maghsoudi} is a Ph.D. student majoring in the Department of Bioengineering at Temple University. He has been a research assistant in the Spence lab for three years. He got his MS degree in medical radiation engineering from Shahid Beheshti University and his BS degree in biomedical engineering (with electrical engineering minor) from Isfahan University. His research interests include image and signal processing, computer vision, neuroscience, biomechanics, and medical imaging devices. His current research is focused on developing a software to track landmarks in the body of running rodents and make 3D model of those markers. Author and coauthor of more than 13 papers.\\

\textbf{Annie Vahedipour} received her BS and MS degrees in Mechanical Engineering from North Carolina State University and Southern Methodist University, respectively. She is currently a Ph.D. student at Temple University and works as a research assistant in the Spence lab. Author and coauthor of two papers. Her work focuses on developing and applying new technologies for the application of external and internal manipulation of the nervous system in so called “neuromechanical” perturbations. \\

\textbf{Benjamin D. Robertson} is a postdoctoral researcher in the Spence lab in the Temple University Department of Bioengineering.  He completed his BS degree in Applied Physics at Emory University, and his PhD in the Joint Department of Biomedical Engineering and UNC-Chapel Hill and NC State University.  Author and coauthor of more than 11 papers. His current research is focused on the role of sensory systems in modulating recovery from Spinal Cord Injury. \\

\textbf{Andrew J. Spence} is an Associate Professor in the Department of Bioengineering at Temple University. He got his PhD from  Cornell University. His research is focused on understanding the control and biomechanics of movement, through an integrative and multidisciplinary approach that combines biology, engineering, mathematics, and molecular genetic tools. Applications of this work are found in spinal cord injury, rehabilitation, neuromuscular disease, and prosthetics, as well as in bio-inspired robotics. Author and coauthor of more than 75 papers.
\end{document}